\documentclass[acmlarge]{acmart}
\usepackage{bm}
\usepackage{subfigure}
\usepackage[vlined,ruled,linesnumbered]{algorithm2e}
\usepackage{multirow}
\usepackage{enumitem}

\usepackage{hyperref}
\hypersetup{
    colorlinks=true,
    linkcolor=blue,
    filecolor=magenta,
    urlcolor=blue,
}

\AtBeginDocument{%
  }

\setcopyright{acmlicensed}
\acmJournal{IMWUT}
\acmYear{2025} \acmVolume{9} \acmNumber{4} \acmArticle{191} \acmMonth{12}\acmDOI{10.1145/3770707}

\acmISBN{XXX-X-XXXX-XXXX-X/XX/XX}

\author{\href{https://orcid.org/0000-0002-5073-5561}{Yunzhe Li}}
\affiliation{
  \institution{Shanghai Jiao Tong University}
  \city{Shanghai}
  \country{China}
}
\email{yunzhe.li@sjtu.edu.cn}

\author{\href{https://orcid.org/0009-0000-6279-6540}{Jiajun Yan}}
\affiliation{
  \institution{Shanghai Jiao Tong University}
  \city{Shanghai}
  \country{China}
}
\email{yanjiajun@sjtu.edu.cn}

\author{\href{https://orcid.org/0009-0008-2850-7643}{Yuzhou Wei}}
\affiliation{
  \institution{Shanghai Jiao Tong University}
  \city{Shanghai}
  \country{China}
}
\email{1340868813@sjtu.edu.cn}

\author{\href{https://orcid.org/0009-0001-1047-3317}{Kechen Liu}}
\affiliation{
  \institution{Columbia University}
  \city{New York}
  \country{USA}
}
\email{kl3469@columbia.edu}

\author{\href{https://orcid.org/0009-0007-4629-7769}{Yize Zhao}}
\affiliation{%
  \institution{University of Electronic Science and Technology of China}
  \city{Chengdu}
  \country{China}
}
\email{202322080612@std.uestc.edu.cn}

\author{\href{https://orcid.org/0000-0001-8857-0144}{Chong Zhang}}
\affiliation{%
  \institution{Southwest Petroleum University}
  \city{Chengdu}
  \country{China}
}
\email{zhangchong92@swpu.edu.cn}

\author{\href{https://orcid.org/0000-0001-8657-5064}{Hongzi Zhu}}
\affiliation{
  \institution{Shanghai Jiao Tong University}
  \city{Shanghai}
  \country{China}
}
\email{hongzi@sjtu.edu.cn}
\authornote{Corresponding author}

\author{\href{https://orcid.org/0000-0002-4361-012X}{Li Lu}}
\affiliation{%
  \institution{University of Electronic Science and Technology of China}
  \city{Chengdu}
  \country{China}
}
\email{luli2009@uestc.edu.cn}

\author{\href{https://orcid.org/0000-0002-5253-2549}{Shan Chang}}
\affiliation{%
  \institution{Donghua University}
  \city{Shanghai}
  \country{China}
}
\email{changshan@dhu.edu.cn}

\author{\href{https://orcid.org/0000-0003-0034-2302}{Minyi Guo}}
\affiliation{
  \institution{Shanghai Jiao Tong University}
  \city{Shanghai}
  \country{China}
}
\email{guo-my@cs.sjtu.edu.cn}

\renewcommand{\shortauthors}{Li et al.}

\begin{document}

\title{BlinkBud: Detecting Hazards from Behind via Sampled Monocular 3D Detection on a Single Earbud}

\begin{abstract}
Failing to be aware of speeding vehicles approaching from behind poses a huge threat to the road safety of pedestrians and cyclists. In this paper, we propose BlinkBud, which utilizes a single earbud and a paired phone to online detect hazardous objects approaching from behind of a user. The core idea is to accurately track visually identified objects utilizing a small number of sampled camera images taken from the earbud. To minimize the power consumption of the earbud and the phone while guaranteeing the best tracking accuracy, a novel 3D object tracking algorithm is devised, integrating both a Kalman filter based trajectory estimation scheme and an optimal image sampling strategy based on reinforcement learning. Moreover, the impact of constant user head movements on the tracking accuracy is significantly eliminated by leveraging the estimated pitch and yaw angles to correct the object depth estimation and align the camera coordinate system to the user's body coordinate system, respectively. We implement a prototype BlinkBud system and conduct extensive real-world experiments. Results show that BlinkBud is lightweight with ultra-low mean power consumptions of 29.8 mW and 702.6 mW on the earbud and smartphone, respectively, and can accurately detect hazards with a low average false positive ratio (FPR) and false negative ratio (FNR) of 4.90\% and 1.47\%, respectively. 
\end{abstract}

\ccsdesc[500]{Human-centered computing~Ubiquitous and mobile computing}

\keywords{Hazardous object detection, monocular 3D object tracking, wearable devices, collision warning}


\maketitle

\section{INTRODUCTION}
The negligence and distraction of pedestrians and cyclists have turned into the main contributing factors to related traffic crashes. For instance, as reported by the National Highway Transportation Safety Administration (NHTSA), the 7,522 pedestrians and 1,105 bicyclists were killed in the U.S. in 2022 \cite{nhtsa2024pedestrians}. It is of great danger to use smartphones or personal audio devices, \emph{e.g.,} noise-canceling earbuds or headphones, while walking or bicycling, as this can increase the risk of collision with vehicles or cyclists \cite{yadav2022systematic, kegalle2025watch, krishna2025does}. The situation gets even more trickier with an ever increasing number of electric vehicles, which are extremely quiet and accelerate fast, making them hard to perceive their approach. As a result, it is of great significance if mobile devices such as earbuds can detect vehicles approaching from the behind and alert users to avoid these hazards.
 

A practical solution for detecting hazardous objects approaching from behind must achieve high detection accuracy in complex and dynamic traffic scenarios, operate efficiently on resource-constrained mobile devices, and provide sufficiently low response latency to enable timely alerts. However, existing approaches fall short of meeting these requirements. Infrastructure-based methods rely on roadside sensors such as cameras \cite{kegalle2025watch, ka2019study, vladyko2020method} and LiDAR \cite{shi2022vips, he2023vi} to detect nearby vehicles and issue warnings, but they are limited by high deployment costs and sparse coverage. Wearables-based methods either exchange location data with vehicles via wireless communication (e.g., Bluetooth \cite{gelbal2022mobile} and 5G \cite{aleva2024augmented}) or detect acoustic signals generated by vehicles \cite{lee2011acoustic, conter2014austrian}. While wireless approaches depend on the availability of compatible modules in vehicles, acoustic-based methods are increasingly ineffective due to the quiet operation of electric vehicles. Consequently, none of these approaches can simultaneously ensure accuracy, responsiveness, and lightweight operation for detecting hazardous objects approaching from behind.


In this paper, we propose a novel system, \emph{BlinkBud}, which utilizes a novel earbud and a paired smartphone to online detect hazardous objects approaching from the behind of a user. In BlinkBud, we define a \emph{blink} as the process of capturing an image via the earbud camera and analyzing it on the paired smartphone for 3D object detection. Rather than continuously streaming video, BlinkBud performs blinks intermittently to reduce power and computational overhead. Foreseeing the trend toward integrating cameras into earbuds to enhance environmental perception, as suggested by ongoing developments such as Apple's AirPods with cameras \cite{9to5mac2025airpods}, Samsung's Galaxy Buds with cameras \cite{samsung2025buds}, and Meta's Camerabuds \cite{meta2024camerabuds}, the core idea of BlinkBud is to utilize a camera integrated into an earbud to collect on-demand images behind the user. Moreover, the phone is employed to accurately track each approaching objects, based on the instantaneous 3D object detection results obtained on a small number of \emph{sampled} camera images. When the estimated time for an object to hit the user is less than a predefined safe threshold, both the phone and the earbud set off an alarm to the user. 

The BlinkBud design faces two main challenges as follows.  
First, power consumption is essential to both the earbud and the phone as they are battery powered. How to minimize the computational and communication overhead while guaranteeing the accuracy of 3D object tracking is of great challenge. As the motion equation of a tracked vehicle can be easily estimated based on previous 3D object detection results, it is straightforward to adopt Kalman filter to estimate future locations of the vehicle and effectively reduce the number of required blinks. However, due to the non-negligible 3D object detection errors, particularly for far objects, and the sudden motion state changes of tracked objects, the confidence of Kalman filter estimation drops significantly over time. To tackle this challenge, we propose an reinforcement-learning-based optimal blink sampling strategy to online characterize the complex relationship among the distance of the object,  the confidence of Kalman filter estimation, and the elapsed time since last blink, and make the optimal decision about when to take a new blink. As a result, high object tracking accuracy can be achieved with the minimum number of required blinks required. 
In addition, to further reduce power consumption, we use grayscale images to shrink the size of the images to be transmitted, and adopt a lightweight 3D object detection scheme based on a compressed 2D object detection neural network combined with pinhole-camera-model-based depth estimation.

Second, the head movements of a user cause constant changes of the extrinsic camera parameters, posing a significant challenge to the object tracking accuracy. 
On one hand, existing monocular depth estimation methods are highly sensitive to the height of pixels in an image, which varies significantly with the pitch rotation of the user's head when looking up and down. On the other hand, the location of the same object in the camera coordination system drastically jumps before and after a yaw rotation of the user's head when looking left and right, which makes objects hard to track.
To deal with this challenge, we amount an inertial measurement unit (IMU) to estimate the pitch and yaw rotation angles. Then, the estimated pitch angle is used to correct the object depth estimation and the estimated yaw angle is used to align the camera coordinate system to the user's body coordinate system, deriving high and robust object tracking accuracy in the condition of normal head movements.

We implement a prototype BlinkBud system on a customized earbud and three representative smartphones, respectively. Specifically, the earbud mainly consists of a mini ESP32-S3 MCU evaluation board, a printed circuit board (PCB) with a 6-axis IMU module and a low voltage OV2640 color CMOS UXGA camera, and a 3D-printed shell. In this implementation, we utilize 2.4 GHz Wi-Fi integrated in the MCU to exchange data between the earbud and a smartphone. 
To evaluate BlinkBud, we establish a comprehensive real-world vehicle approaching dataset, involving 262 video clips of one minute long and the corresponding head movement IMU readings collected in various scenarios with respect to different volunteer users, transportation modes of users, road types with varying numbers and types of vehicles, and light conditions. 
We conduct extensive hazard detection experiments\footnote{All experiments were conducted with Institutional Review Board (IRB) approval from the host university.}. Results show that BlinkBud is lightweight and agile to detect hazards accurately with ultra low power consumptions of 29.8 mW and 702.6 mW on the earbud and smartphone, respectively. Furthermore, BlinkBud can achieve a low average false positive ratio (FPR) of 4.90\% and a low average false negative ratio (FNR) of 1.47\%. Moreover, positive user feedback from 43 volunteers demonstrates both the efficacy and the strong user acceptance of the BlinkBud design.

We highlight the main contributions made in this paper as follows:

\begin{itemize}
	\item We have devised a robust and lightweight 3D object detection scheme, which can obtain accurate 3D coordinates of objects from 2D images even when users are in different transportation modes and constantly move their heads;
	\item We have proposed an effective 3D object tracking algorithm by integrating a Kalman filter based trajectory estimation scheme and an optimal image sampling strategy, which can minimize the power consumption of the system while guaranteeing the best tracking accuracy;
	\item We have implemented a prototype system and validated the effectiveness and feasibility of the BlinkBud system based on extensive real-world experiments.
\end{itemize}

\section{RELATED WORK}

\subsection{Approaching Objects Detection} 

Approaching object detection is a critical task for applications such as pedestrian safety, autonomous driving, and assistive technologies. A variety of sensor modalities have been explored in the literature to address this problem. Vision-based sensors, including monocular or stereo cameras \cite{hakkert2002evaluation, vladyko2020method, kegalle2025watch} and LiDAR systems \cite{shi2022vips, he2023vi, mohapatra2021bevdetnetbirdseyeview}, have been widely adopted due to their ability to provide rich spatial and semantic information, enabling precise localization and classification of approaching objects. However, these approaches typically demand significant computational resources and rely on favorable environmental conditions, such as clear line-of-sight and sufficient lighting, which limits their deployment mainly to roadside infrastructure or vehicles equipped with high-end hardware.
To enable lightweight and energy-efficient detection, acoustic-based methods leveraging microphones have also been investigated \cite{lee2011acoustic, conter2014austrian}. These methods exploit characteristic noise signatures generated by vehicles to infer the presence and direction of approach. While acoustic approaches can be implemented on mobile and wearable devices, their effectiveness is substantially reduced in noisy environments and notably in detecting electric vehicles that produce minimal sound \cite{Pardo-Ferreira2020}.
Another promising direction involves cooperative detection through wireless communication technologies, where approaching vehicles and pedestrians exchange location and motion information via protocols such as Bluetooth \cite{gelbal2022mobile} or emerging 5G networks \cite{aleva2024augmented}. Despite the advantages of this strategy in enhancing situational awareness without heavy sensing requirements, its performance heavily depends on widespread adoption of compatible communication modules among vehicles and pedestrians, which remains a significant practical challenge.
Our approach complements these prior methods by focusing on lightweight, on-device sensing that does not rely on external infrastructure or extensive communication, while maintaining privacy and efficiency.

\subsection{Object Detection}

Object detection is a foundational task in computer vision that involves localizing and classifying objects within images. Traditional two-stage detectors \cite{girshick2014rich, ren2017faster} achieve high accuracy by first generating region proposals and subsequently classifying these regions, albeit at relatively high computational cost. In contrast, one-stage detectors \cite{cheng2024yolo, zhao2024ms} streamline the detection pipeline by predicting bounding boxes and class probabilities in a single forward pass, trading off some accuracy for real-time performance.
For deployment on resource-constrained devices, lightweight architectures \cite{sandler2018mobilenetv2, lin2017focal} are proposed, which optimize model size and inference speed while retaining reasonable detection accuracy.
In the realm of 3D object detection, approaches \cite{chabot2017deep, zhou2025exploiting} leverage multi-task learning or prior geometric knowledge to jointly predict 2D and 3D bounding boxes. Nevertheless, the computational complexity and memory requirements of these models present challenges for real-time operation on mobile and embedded platforms.
Our proposed BlinkBud bridges these gaps by integrating efficient 2D detection with lightweight 3D estimation, enabling real-time performance under constrained computational budgets.

\subsection{Monocular Depth Estimation}

Monocular depth estimation, the task of inferring depth information from a single RGB image, plays a vital role in understanding scene geometry and is especially important for positioning vehicles relative to their surroundings. Early supervised methods \cite{eigen2014depth} employ multi-scale convolutional networks to regress depth maps directly from RGB images, laying foundational work for learning-based scene geometry understanding. However, such methods require dense ground-truth depth labels, which are often expensive to obtain. To address this, the self-supervised approach \cite{godard2019digging} is proposed to leverage photometric consistency across stereo pairs or consecutive frames, eliminating the need for explicit depth supervision and enabling broader scalability. Building on this direction, MoGDE \cite{zhou2025exploiting} introduces a domain-specific enhancement by estimating ground plane depth to improve monocular 3D object detection on mobile platforms, which is especially effective in driving and robotics contexts where ground geometry is informative. Complementing these vision-based methods, a mmWave radar-based solution \cite{feng20243d} is proposed, which estimates 3D bounding boxes by scanning with commercial off-the-shelf (COTS) radar modules. This radar-centric design mitigates visual degradation under low-light or occluded conditions and introduces an alternative sensing modality for depth estimation. Together, these works illustrate the evolution from supervised to self-supervised and multimodal approaches, aiming to improve the generalization, robustness, and deployability of depth perception systems.
However, many existing methods entail high computational cost and latency, which limits their suitability for real-time mobile applications.
Our method prioritizes computational efficiency and robustness by leveraging geometric cues, facilitating accurate depth estimation suitable for on-device processing.

\subsection{Multi-object Tracking}

Recent advances in multi-object tracking (MOT) have moved beyond the traditional tracking-by-detection pipeline toward more unified or hybrid designs that tightly couple detection and association. OmniTracker \cite{wang2025omnitracker} introduces a general tracking-with-detection framework that adaptively generates anchors for integrated detection and tracking, enabling end-to-end optimization and robust long-term performance. In the 3D domain, AB3DMOT \cite{Weng2020_AB3DMOT} establish a modular baseline and propose the HOTA metric to jointly evaluate detection and association accuracy, offering a more comprehensive benchmark for 3D MOT. Hybrid-SORT \cite{yang2024hybrid} enhances the classical SORT algorithm by incorporating weak visual cues and transition learning, demonstrating that even lightweight semantic and motion information can yield improved tracking robustness—though it remains a two-stage, post-detection method. In contrast, NetTrack \cite{zheng2024nettrack} presents a fully convolutional network that jointly handles detection and association for fast-moving or deformable targets, eliminating explicit matching heuristics. Transformer-based models have also gained traction for their ability to model long-range dependencies: TrackFormer \cite{meinhardt2022trackformer} frames MOT as a set prediction task using persistent queries across frames, while Global Tracking Transformers \cite{zhou2022global} apply global attention to improve association under occlusion and clutter. These approaches collectively reflect a shift toward tighter integration of spatial, temporal, and appearance cues within unified tracking architectures.
Despite advances, achieving efficient and accurate MOT on mobile or wearable devices remains challenging due to limited computational resources and the additional complexity introduced by head movements in wearable camera settings \cite{jo2024watchcap}. This paper contributes to this area by proposing lightweight tracking techniques that can mitigate the impact of head movements, thereby improving robustness in egocentric scenarios.

\section{PROBLEM DEFINITION}

\subsection{System Model}

We consider three types of entities in the system:

\begin{itemize}
	
	\item \textbf{Users}: We consider BlinkBud users to be pedestrians who may stand, walk, or jog in environments with motor traffic, as well as on sidewalks that are commonly shared with bicycles. A user needs to properly wear the earbud with the camera upright facing the back of the user and is allowed to move his/her head at will. 
	\item \textbf{Earbud}: The earbud is equipped with a camera and an IMU. It communicates with the smartphone of a user via a wireless link such as WiFi or Bluetooth. Upon instructions received from the phone, it can take an image and record acceleration and gyroscope readings which are sent back to the phone for processing. It can also play alarm sounds when the phone indicates a detected hazard.
	\item \textbf{Smartphone}: The smartphone has certain computing and communication capabilities. It makes the optimal blink scheduling and conducts 3D object tracking based on results of existing blinks. Moreover, it assesses the potential risk level of the user being hit and triggers alarms accordingly. To address privacy concerns, all image processing must be conducted locally on the smartphone without permitting any form of data storage. Furthermore, the processing of images captured in indoor environments is automatically disabled upon detection by the device. This indoor detection is reliably determined based on the presence of specific objects identified within the images.
\end{itemize}

\begin{figure}
 \centering
    \includegraphics[width=4cm]{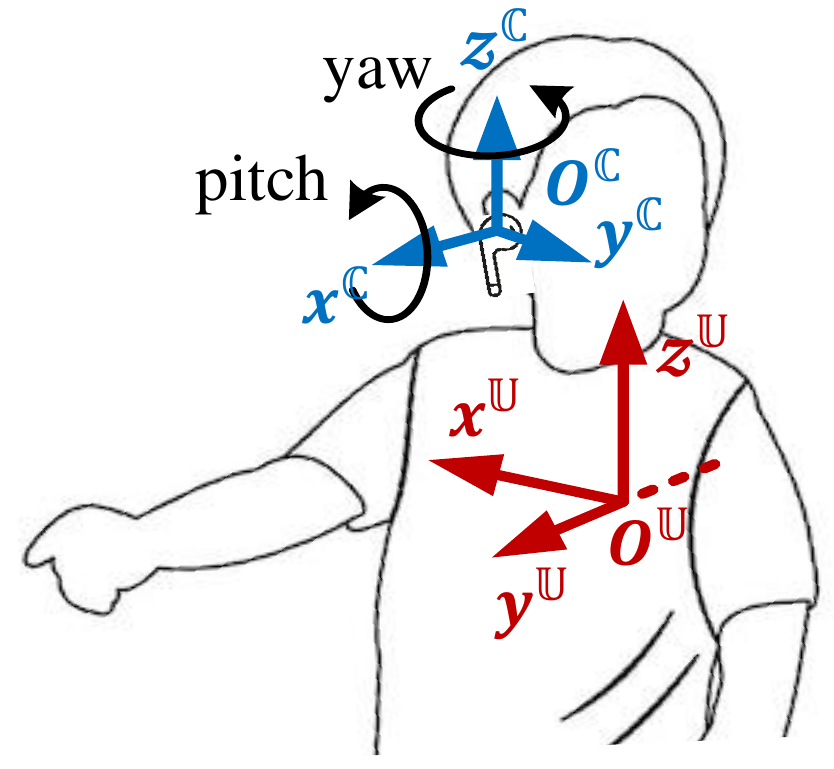}
    \caption{The illustration of the camera coordinate system on the earbud and the user coordinate system.}
    \label{fig:coordinate}
\end{figure}

\subsection{Rear Hazard Detection Problem}

As illustrated in Figure \ref{fig:coordinate}, we denote the coordinate system of a user as $\mathbb{U}$, that originates at the user's heart $O^\mathbb{U}$ with its x-, y- and z-axes denoted as $x^\mathbb{U}, y^\mathbb{U}, z^\mathbb{U}$ pointing to the right side, the front, and the above of the user, respectively. In addition, we denote the coordinate system of the camera on the earbus as $\mathbb{C}$, which aligns with that of the head of the user. Note that the displacement between the camera coordinate system and the user coordinate system shown in Figure 1 is only for illustration purposes and has negligible impact on the design of BlinkBud. We consider two types of head movements, \emph{i.e.}, yaw rotation when looking around and pitch rotation when looking up and down, which also cause the same camera rotations.


Denote the moving objects behind the user as $o_1, o_2, \cdots, o_N$, where $N$ is the total number of identified objects. The 3D position of object $o_i$ in the camera coordinate is denoted as $o^\mathbb{C}_i = \{x^\mathbb{C}_i, y^\mathbb{C}_i, z^\mathbb{C}_i\}$, and its position in the user coordinate is $o^\mathbb{U}_i = \{x^\mathbb{U}_i, y^\mathbb{U}_i, z^\mathbb{U}_i\}$, where $i \in [1, N]$. The \emph{Rear Hazard Detection (RHD)} problem is defined as to minimize the number of required blinks while ensuring accurate tracking for all objects within a certain distance from the user:
\begin{equation*}
\min \sum_{t=1}^T A_t,
\end{equation*}
subject to:
\begin{equation*}
    \sum_{i=1}^N \mathcal{E}(o^\mathbb{U}_i, \hat{o}^\mathbb{U}_i) < \varepsilon,  \forall \, o^\mathbb{U}_i \, \text{such that} \, \|o^\mathbb{U}_i\| \leq d_{\text{max}},
\end{equation*}
where $A_t$ indicates whether a blink should be conducted at time $t$, \emph{i.e.}, $A_t = 1$ if yes and $A_t = 0$ otherwise; $\mathcal{E}(o^\mathbb{U}_i, \hat{o}^\mathbb{U}_i)$ represents the tracking error between the actual position $o^\mathbb{U}_i$ and predicted position  $\hat{o}^\mathbb{U}_i$ of object $o_i$ in the user coordinate system; $\varepsilon$ is the preset accuracy threshold;
$d_{\text{max}}$ is the maximum distance within which objects are subjected to the accuracy constraint;  $\|o^\mathbb{U}_i\|$ denotes the distance of object $o_i$ from the user.

\section{DESIGN OF BLINKBUD}
\subsection{Overview}

\begin{figure}

    \centering
    \includegraphics[width=0.5\linewidth]{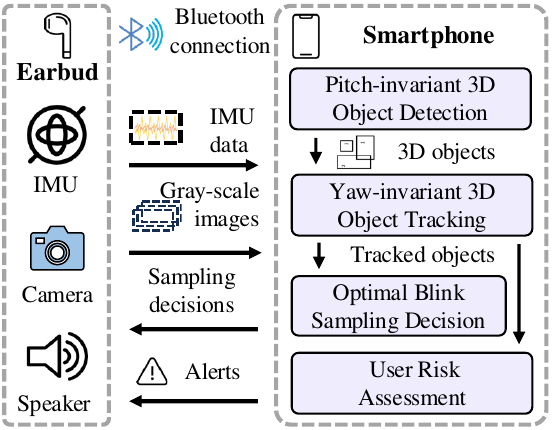}
    \caption{The architecture of BlinkBud, where the earbud and the phone cooperate to track hazardous moving objects from the behind of a user with the minimal number of instantaneous visual perceptions.}
    \label{fig:overview}
\end{figure}

As illustrated in Figure \ref{fig:overview}, BlinkBud utilizes a earbud and a paired phone to accurately track hazardous moving objects from the behind of a user at the minimal computational and communication costs. To this end, BlinkBud effectively integrates the following four components.

\subsubsection{Pitch-invariant 3D Object Detection} 
Upon receiving a grayscale image from the earbud and the corresponding IMU data, the phone first conducts 2D object detection on that image. Then, the pitch rotation angle of the user's head is estimated to accurately estimate the depth information of each identified object based on the pinhole camera model. As a result, given the estimated depth information, 3D coordinates of those identified 2D objects in the camera coordinate system are estimated in a pitch-invariant way.
 
\subsubsection{Yaw-invariant 3D Object Tracking} 
First, the yaw rotation angle of the user's head is estimated to convert the estimated locations of objects from the camera coordinate system into the user coordinate system. Then, as the motion equation of a tracked vehicle can be easily estimated based on previous 3D object detection results, Kalman filter is conducted on the phone to estimate future locations of identified objects. 

\subsubsection{Optimal Blink Sampling Decision}

To overcome the dropping confidence of Kalman filter estimation due to the non-negligible 3D object detection errors, an reinforcement-learning-based optimal blink sampling strategy is conducted on the phone to online characterize the complex relationship among the distance of the object,  the confidence of Kalman filter estimation, and the elapsed time since last blink, and make the optimal decision about when to take a new blink. Consequently, an instruction is sent to the earbud to take a new image and IMU data.

\subsubsection{User Risk Assessment}
Based on estimated trajectories of tracked objects, a risk level is assessed by analyzing the estimated time-to-collision (TTC) of each object. If the risk level exceeds a certain threshold, the phone will trigger the earbud to play warning sound, alerting the user to this potential hazard.

\subsection{Pitch-invariant 3D Object Detection}



\begin{figure}[]
	\centering
	\subfigure[A car can be first detected over 12 meters at 640p.]{
		\begin{minipage}[b]{0.46\linewidth}
			\centering
			\includegraphics[height=4.2cm]{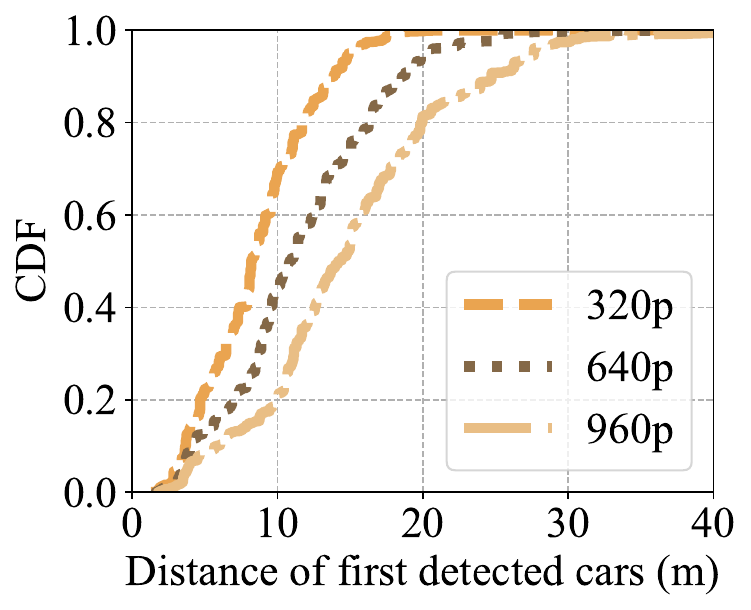}
		\end{minipage}
		\label{fig:first_detect_car}
	}
	\subfigure[A cycle can be first detected over 6 meters at 640p.]{
		\begin{minipage}[b]{0.46\linewidth}
			\centering
			\includegraphics[height=4.2cm]{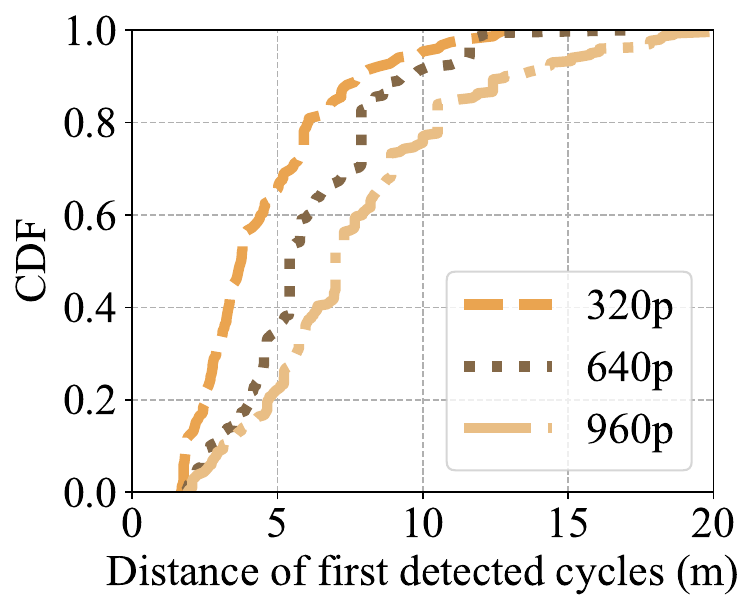}
		\end{minipage}
		\label{fig:first_detect_cyclist}
	}
		
	\caption{CDF of distances when cars and cycles are first detected using images of different resolutions.}
	\label{fig:emp_analysis}
\end{figure}

\subsubsection{Lightweight 2D Object Detection}
Given each captured grayscale image $I$, a lightweight 2D object detector, denoted as $\mathcal{M}_{2D}$, is utilized to detect 2D objects $o^{\mathbb{I}}_1, o^{\mathbb{I}}_2, \cdots, o^{\mathbb{I}}_{n}$ in the image $I$:
\begin{equation*}
    o^{\mathbb{I}}_1, o^{\mathbb{I}}_2, \cdots, o^{\mathbb{I}}_{n} = \mathcal{M}_{2D}(I)
\end{equation*}

We adopt grayscale images as input for object detection based on empirical analysis conducted using our self-collected rear-approach dataset. This dataset comprises synchronized grayscale images and LiDAR-based ground-truth annotations, collected via an OV6420 grayscale camera embedded in an earbud and a 32-line RoboSense Helios LiDAR mounted on a tripod. Details of the dataset are provided in Subsection~\ref{sec:dataset}.
In a representative evaluation on a randomly selected video sequence, grayscale input yielded a detection accuracy of 95.83\%, compared to 94.45\% using RGB input. We attribute this improvement to the nature of the rear-approach detection task, which relies primarily on structural cues such as object boundaries and contours, rather than color information. Grayscale images reduce variability caused by illumination and color shifts, thereby enhancing robustness. Additionally, the use of grayscale significantly lowers computational overhead and energy consumption, which are factors that are critical for resource-constrained wearable systems. For these reasons, we adopt grayscale input as the default configuration in our detection pipeline.

Figure~\ref{fig:first_detect_car} and Figure~\ref{fig:first_detect_cyclist} show the cumulative distribution functions (CDFs) of the initial detection distances for rear-approaching objects, including cars and cycles (i.e., bicycles, electric bicycles, and motorcycles). The evaluation is performed using a YOLOv5s-based object detection model pre-trained on the KITTI dataset and tested on our self-collected rear-approach video clips. Three input resolutions, \emph{i.e.}, 320p, 640p, and 960p, are compared to assess the trade-off between detection range and computational cost.
The results indicate that higher input resolutions yield longer detection distances, which is beneficial for early hazard awareness. However, the use of higher resolutions also incurs significantly greater computational and communication overhead, leading to increased power consumption on both the earbud and the paired smartphone. This introduces a practical constraint in resource-constrained wearable systems. As such, it is crucial to determine the lowest resolution that achieves acceptable detection performance while minimizing energy and processing demands. Our subsequent analysis adopts 640p as a balanced configuration, offering a favorable trade-off between accuracy and system efficiency.
We can see that the average distance of cars and cyclists with a detection resolution of 640p is 12 meters and 6 meters, respectively. In contrast, the average human reaction time to auditory signals is 0.5 seconds~\cite{grice1982human}. 
This time corresponds to a travel distance of 4.1 meters for cars at 30 km/h and 2.8 meters for cyclists at 20 km/h. As a result, a lightweight YOLOv5s-based 2D object detection model with an image resolution of 640p is sufficient. 

\subsubsection{Object Depth Estimation}
For each $o^{\mathbb{I}}$ in the detected 2D objects $\{o^{\mathbb{I}}_1, o^{\mathbb{I}}_2, \cdots, o^{\mathbb{I}}_{n} \}$, the depth $z^{\mathbb{C}}$ of $o^{\mathbb{I}}$ is then estimated using the camera’s intrinsic parameters and the pitch angle provided by the gyroscope. The core idea is to calculate depth based on the vertical position of a vehicle’s projection relative to the horizon in the image plane. The horizon serves as a reference line, and depth is determined using the pixel deviation between the vehicle’s bottom contact point with the ground and the horizon. Details are provided in Appendix \ref{sec:depth}.

\subsubsection{2D to 3D Coordinate Transformation} 

Each detected 2D object is projected to 3D camera coordinates using its estimated depth and the known camera intrinsics. The full transformation is provided in Appendix \ref{sec:trans}.

\begin{figure}
    \centering
    \includegraphics[width=0.35\linewidth]{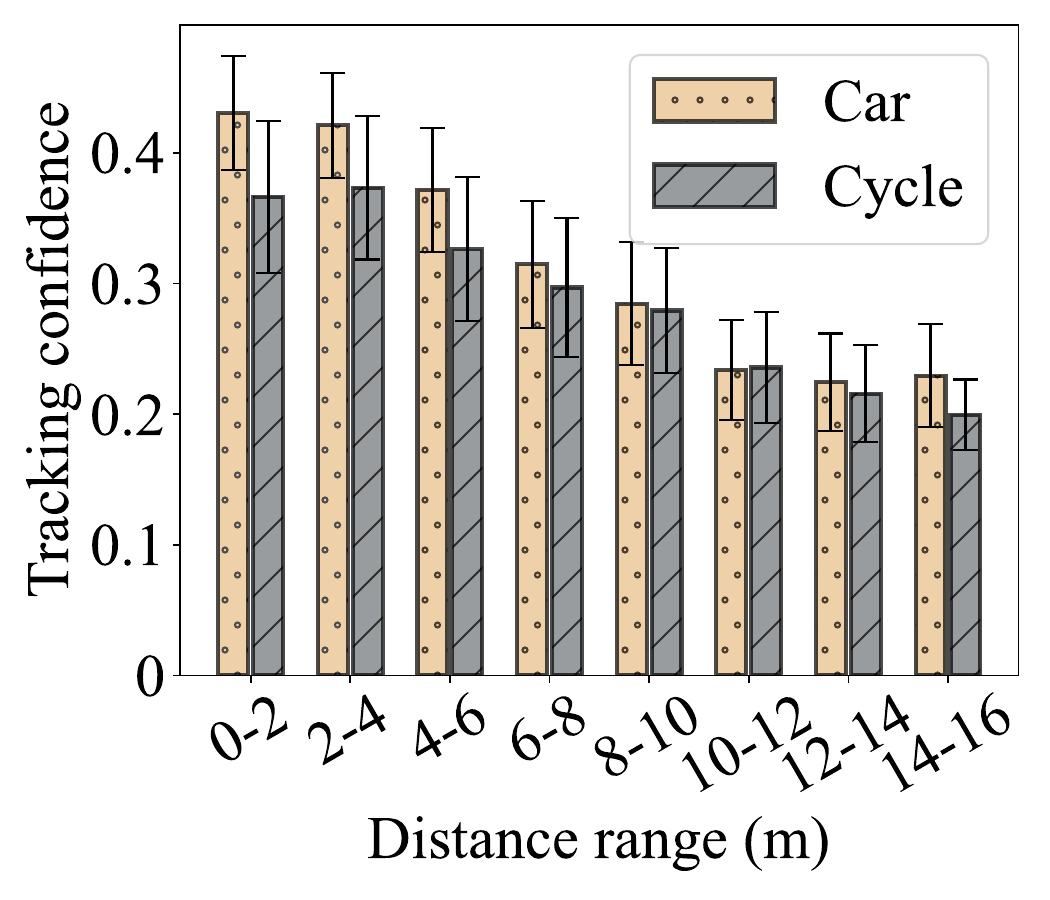}
    \caption{Histogram 
 of tracking confidence with different distances, where the tracking confidence decreases as distance increases.}
    \label{fig:confidence}
\end{figure}

\subsection{Yaw-invariant 3D Object Tracking}

\subsubsection{Multi-object Matching}

Given the sample 3D objects in the user system, BlinkBud needs to match the 3D boxes across the frames. The Kuhn-Munkres algorithm \cite{zhu2016solving} is employed for matching detected objects across frames.
Specifically, the object matching process is formulated as a weighted bipartite graph. In this graph, the target boxes and detection boxes are represented as two disjoint sets of nodes. Each pair of target and detection boxes is connected by an edge, with the weight of the edge reflecting the degree of similarity between the two boxes. BlinkBud seeks to maximize the total correlation between objects across frames, which is quantified by the Intersection over Union (IoU) metric. The Kuhn-Munkres algorithm is then applied to solve the assignment problem.
For detection boxes that are not matched with any existing target, new targets are introduced and added to the object queue. Targets that remain unmatched for a certain number of frames have their unmatched count incremented. Once this count surpasses a predefined threshold, the target is removed from the queue.




\label{sec:EKF}
\subsubsection{Multi-object Tracking}

To track surrounding objects across frames, we transform their 3D positions from the camera to the user coordinate system using the yaw angle derived from IMU data. We model each object’s motion as linear and apply an Extended Kalman Filter (EKF) \cite{khodarahmi2023review} to estimate its position and velocity over time. The EKF accounts for both motion prediction and observation updates, allowing us to maintain accurate trajectories even under partial occlusion or intermittent detection. Full state definitions and equations are provided in Appendix \ref{sec:mot}.

During the tracking based on EKF, the confidence of Kalman filtering at each time step can be computed as the reciprocal of trace the covariance matrix $ \bm{P}$ \cite{yoon2008new}:
\begin{equation*}
    c=\frac{1}{tr(\bm{P})+\gamma},
\end{equation*}
where $\gamma$ denotes a very small constant to avoid dividing by zero.
We perform target tracking based on EFK frame by frame and calculate the distribution of tracking confidence for objects at different distances on the clips from our self-collected dataset, as shown in Figure \ref{fig:confidence}.
We can see that despite frame-by-frame sampling, tracking confidence still decreases as distance increases. This is because the target detector cannot accurately track distant objects, considering its limited capabilities.
However, the danger of distant objects is much smaller than that of nearby objects, and we do not need to track them accurately before they arrive. Relying solely on confidence to decide whether to sample will make the tracking model overly focused on distant objects, which is in fact a waste. We need to comprehensively consider the impact of object distance and tracking confidence on blink sampling decisions to make the optimal sampling decision.

\subsection{Optimal Blink Sampling Decision}

To reduce unnecessary high-rate sampling and power consumption, we propose an adaptive frame sampling strategy based on online reinforcement learning. Specifically, we adopt the SARSA algorithm to learn when to sample frames by modeling the problem as a Markov Decision Process (MDP). The state captures system uncertainty, object proximity, and time since last sample. The action determines whether to sample or skip the frame. The reward function balances tracking confidence improvement against sampling cost.
Using an $\epsilon$-greedy policy with decaying exploration and a reward-driven update rule, the agent gradually learns an efficient sampling policy that adapts to dynamic hazards in real time.
Details of the state space, reward function, and value update are provided in Appendix \ref{sec:blink_sampling}.

\begin{theorem}
\label{theo}
The adaptive frame sampling strategy converges to the optimal policy.
\end{theorem}

The proof of Theorem \ref{theo} is provided in Appendix \ref{sec:proof}.

\subsection{User Risk Assessment}

Given the tracked trajectories $r^{\mathbb{U}}_1, r^{\mathbb{U}}_2, \cdots, r^{\mathbb{U}}_{n}$, BlinkBud performs a risk assessment to evaluate the potential danger posed by each object, based on the time to collision. Specifically, denote an object as $o_t^{\mathbb{U}}$ at time $t$, the motion state of $o_t^{\mathbb{U}}$ as $[x_t^\mathbb{U}, z_t^\mathbb{U}, \dot{x}_t^\mathbb{U}, \dot{z}_t^\mathbb{U}]$.
The radial time to collision can be calculated as follows:
\begin{equation*}
    t^{c} = -\frac{{x_t^\mathbb{U}}^2 + {z_t^\mathbb{U}}^2}{x_t^\mathbb{U} \dot{x}_t^\mathbb{U} + z_t^\mathbb{U} \dot{z}_t^\mathbb{U}}.
\end{equation*}
Note that when object $o_t^{\mathbb{U}}$ approaches the person radially, $t^{c} > 0$; when the vehicle moves away radially, $t^{c} < 0$. 
The risk level of objects can then be defined as follows:
\begin{equation*}
    \kappa = \max (0, 1 - \frac{t^{c}}{t^{r}}),
\end{equation*}
where $t^{r}$ denotes a user-configurable risk sensitivity threshold. For each object at time $t$, the corresponding risk level can be calculated. The overall risk level at time $t$, denoted as $\Gamma$, is defined as the maximum of the risk levels of current objects. If $\Gamma$ exceeds a certain threshold, BlinkBud will alert the user to be cautious. The threshold parameter can be set according to the personal safety requirement of each individual user.

\section{PROTOTYPE IMPLEMENTATION}



\begin{figure}[]
    \centering
    \includegraphics[width=0.65\linewidth]{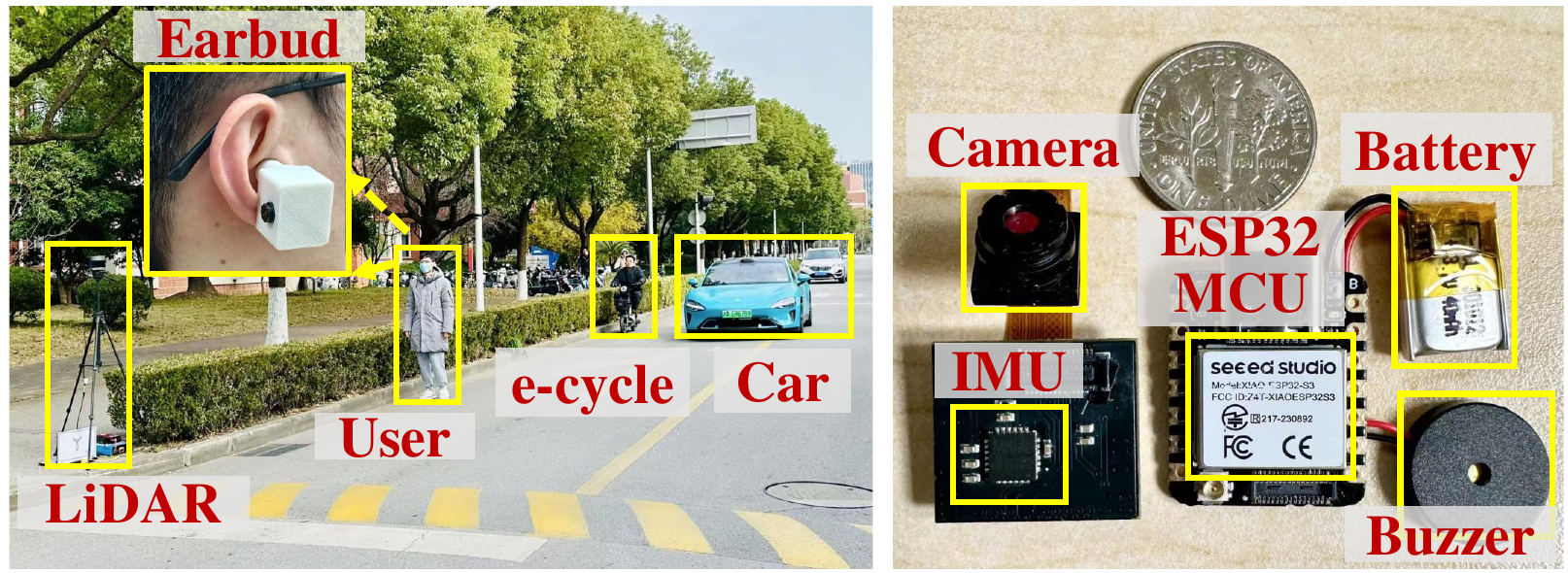}
    \caption{The Prototype implementation of BlinkBud on an earbud (right) and an example of field study (left), where a volunteer wearing the earbud is walking along a road with motor traffic.}
    \label{fig:earbud}
\end{figure}

We implement a prototype system for BlinkBud based on earbuds and a smartphone, as shown in Figure \ref{fig:earbud}, including two parts, \emph{i.e.}, a camera-equipped earbud for sensing and a smartphone for computing. The prototype costs approximately \$18 USD.

\subsection{Implementation of Earbud}

The implementation of the earbuds is illustrated in Figure \ref{fig:earbud}, which is powered by a 3.7V battery and mainly includes a mini ESP32-S3 evaluation board, a customized printed circuit board (PCB) with a 6-axis I$^2$C MPU6050 IMU module and a low voltage OV2640 color CMOS UXGA camera, and a 3D-printed shell.
Specifically, the camera captures the images at a resolution of $640 \times 640$. When the smartphone requests data, the mini EPS32-S3 reads the grayscale image frame from the camera and gyroscope data from the IMU sensor and sends them to the phone based on the HTTP protocol through WiFi. Upon receiving the alert signal from the phone, the buzzer on the earbud will sound to remind the user.

\subsection{Implementation of Smartphone Application}
A smartphone application is developed to continuously receive and process the data received from the earbud, and display the results.
The YOLOv5s \cite{jocher2022ultralytics} model pretrained on the KITTI dataset, is deployed on the smartphone application based on the NCNN inference library \cite{moolchandani2021accelerating}. The smartphone can display the picture accepted by the camera and mark the target tracker and the risk scale in the picture. For targets whose danger level exceeds a certain threshold, the mobile phone sends a control signal to the earbud to trigger an alarm. We implement the application of BlinkBud on 3 typical mobile phones, with configurations listed in Table \ref{tab:phone}. 

\begin{table}[]
\caption{Three types of mobile phones are considered in the prototype.}
\label{tab:phone}
\centering
\scalebox{1}{
\begin{tabular}{cccc}
\toprule
Device    & SoC & Memory & Disk \\ \hline
Redmi K70 &  Snapdragon 8 Gen 2  &  12 GB  &  512 GB    \\ 
Mi11      &  Snapdragon 888   &   8 GB     &  256 GB    \\ 
Redmi K20 &  Snapdragon 730   &   6 GB     &  128 GB    \\ \bottomrule
\end{tabular}
}
\end{table}

\subsection{Collecting Rear Object Approaching Data}
\label{sec:dataset}
We collect a comprehensive real-world moving object approaching dataset, involving 262 video clips of one minute long and the corresponding head movement IMU readings collected on our campus from May to June 2024. 
A total of 10 volunteers were hired to collect data, including 8 males and 2 females. The volunteers are aged between 20 and 30 years old, with heights ranging from 1.6 meters to 1.9 meters. All experiments involving human participants were approved by the Institutional Review Board (IRB) of the host university and conducted in accordance with applicable institutional and local ethical guidelines.
The OV6420 camera equipped on the earbud is utilized to capture rear images of approaching vehicles at a frame rate of 10 FPS.
A 32-line RoboSense Helios LIDAR is mounted on the tripod and captured at a frame rate of 10 FPS to obtain the ground-truth position of the approaching vehicles.
The volunteers are asked to maintain three different transportation modes, \emph{i.e.}, \emph{standing}, \emph{walking} and \emph{jogging}. 
The volunteers walked in their usual manner, which might include natural head movements such as turning their heads to look at the surroundings. 
The data are collected from two typical road types, \emph{i.e.}, \emph{along roads} and \emph{at intersections}, where vehicles may arrive from two directions and four directions, respectively. The number of vehicles and cycles identified at the same time ranges from one to nine. 

The LiDAR-based 3D object detection repository OpenPCDet \cite{openpcdet2020} and 3D object tracking repository AB3DMOT \cite{Weng2020_AB3DMOT} are utilized to obtain the ground truth of 3D coordinates of vehicles, cycles and volunteer users as well.
The LiDAR coordinate system and the user coordinate system are aligned using the RANSAC algorithm \cite{fischler1981random}.
Both LIDAR and images are aligned based on their own UNIX timestamps.






\section{DISCUSSION}

\subsection{Privacy Concerns}
The privacy concern of our vision detection is another issue.
On one hand, after obtaining the images, our system processes them directly on the local endpoint without saving or uploading them to the cloud. On the other hand, similar to a dashcam, most of the images we capture are in public settings, posing no confidentiality or privacy concerns. 
The point of concern arises only when the user is indoors. Therefore, we are considering having the camera halt video streaming upon detecting that the user is indoors, accompanied by an alert to the user, which reminds him to resume streaming upon returning outdoors.



\subsection{Limited Perceptual Range}

The human binocular field of view spans approximately $200^\circ$, leaving a blind region of about $160^\circ$. A single camera is generally insufficient to cover this entire area due to its limited viewing angle. To mitigate this limitation, wide-angle solutions such as fisheye cameras can be employed. However, fisheye cameras generate images that differ substantially from those captured by conventional cameras, which requires specialized datasets for retraining object detection models. Moreover, even with a fisheye camera providing a $180^\circ$ field of view, blind zones persist due to occlusion from the user’s head. These residual blind spots are not necessarily critical, since natural head movements shift both the visual and camera perspectives, thereby reducing or temporarily eliminating portions of the blind region. In addition, the probability of traffic approaching from certain directions is relatively low, for example from the right-hand side for pedestrians walking on the right. Therefore, maintaining continuous coverage of the entire blind area is not essential.





\subsection{Detecting New Hazard Types}

While our current implementation of BlinkBud focuses on detecting rear-approaching cars and cycles, the underlying detection framework is inherently extensible to a broader range of traffic hazards. This extensibility is enabled by two design principles. First, BlinkBud utilizes a modular object tracking pipeline based on a general-purpose vision model. Second, a lightweight sensor-streaming and inference architecture is designed, which can accommodate additional hazard classes with minimal modification.
To support emerging vehicle types such as e-Scooters, delivery robots, or shared micro-mobility devices, the primary requirement is the availability of appropriately labeled training data. Once sufficient samples of these new hazard types are collected, the existing detection model can be fine-tuned or re-trained using transfer learning techniques. This ensures minimal disruption to the rest of the BlinkBud pipeline.

\subsection{Hazard Detection Latency}

A critical aspect in assessing the safety of our system is the end-to-end latency from hazard occurrence to user notification. This latency can be decomposed into three components: (i) image capture and transmission, defined as the time required for the onboard camera to acquire a frame and transfer it to the smartphone; (ii) inference latency, corresponding to the time needed for the perception module to process the image and generate a risk estimate; and (iii) alert latency, representing the time to generate and deliver a warning through the user interface (e.g., auditory or visual feedback). In our implementation, image capture and transmission typically require approximately 30 ms, inference on the embedded GPU averages 39 ms, and alert delivery incurs less than 3 ms. The resulting total end-to-end latency is therefore about 72 ms, substantially lower than the typical human perception–response time of roughly 500 ms \cite{grice1982human}. This breakdown indicates that the system imposes negligible delay relative to human reaction and can thus provide a meaningful temporal margin for hazard avoidance.


\subsection{Choice of Risk Threshold}

It is important to clarify that the \emph{risk threshold} in our system is not a tunable model hyperparameter but a safety criterion derived from the estimated time-to-collision (TTC). In particular, we map the TTC value to a risk score, where higher scores correspond to shorter remaining times and therefore greater urgency. The threshold represents the minimum risk level at which an alert is triggered. In our implementation, the threshold is set to approximately 3.3 seconds, which aligns with the average human reaction time. Conceptually, selecting a smaller threshold results in earlier alerts but increases the likelihood of false alarms, as surrounding vehicles often adjust their trajectories only at closer distances. Conversely, a larger threshold reduces false alarms but risks generating alerts too late for the user to respond effectively. It should be emphasized that the setting of this threshold also depends on users’ psychological tolerance toward safety risks. Our current implementation adopts an average value suitable for the general population. Developing adaptive mechanisms that adjust the threshold to individual user preferences in order to enhance user satisfaction will be addressed in our future work.

\section{EVALUATION}

\subsection{Methodology}

\subsubsection{Candidate Methods}
We consider the following 5 representative candidate methods.


\begin{itemize}
	\item \textbf{Naive Extended Kalman Filter (EKF)} \cite{khodarahmi2023review} tracks the vehicles frame by frame based on  Extended Kalman Filter, which can obtain accurate alert results at the cost of high system overhead.
	
	\item \textbf{Interval Frame Sampling (Intr.)} samples frames at fixed intervals, typically with a pre-determined interval between each frame.

        \item \textbf{Random Frame Sampling (Rand.)} randomly selects frames for processing to reduce the computational burden.

    	\item \textbf{Confidence-based Frame Sampling (Conf.)} \cite{yoon2008new} uses the confidence of object tracking to decide whether to sample a particular frame. When the confidence in object detection is low, the system samples a frame to obtain accurate tracking results.

    	\item \textbf{BlinkBud w/o IMU} tracks objects solely relying on images without incorporating IMU sensors.
\end{itemize}

\begin{figure}[]
	\centering
	\subfigure[FPR in 3 trans. modes]{
		\begin{minipage}[]{0.46\linewidth} 
			\centering
			\includegraphics[height=3cm]{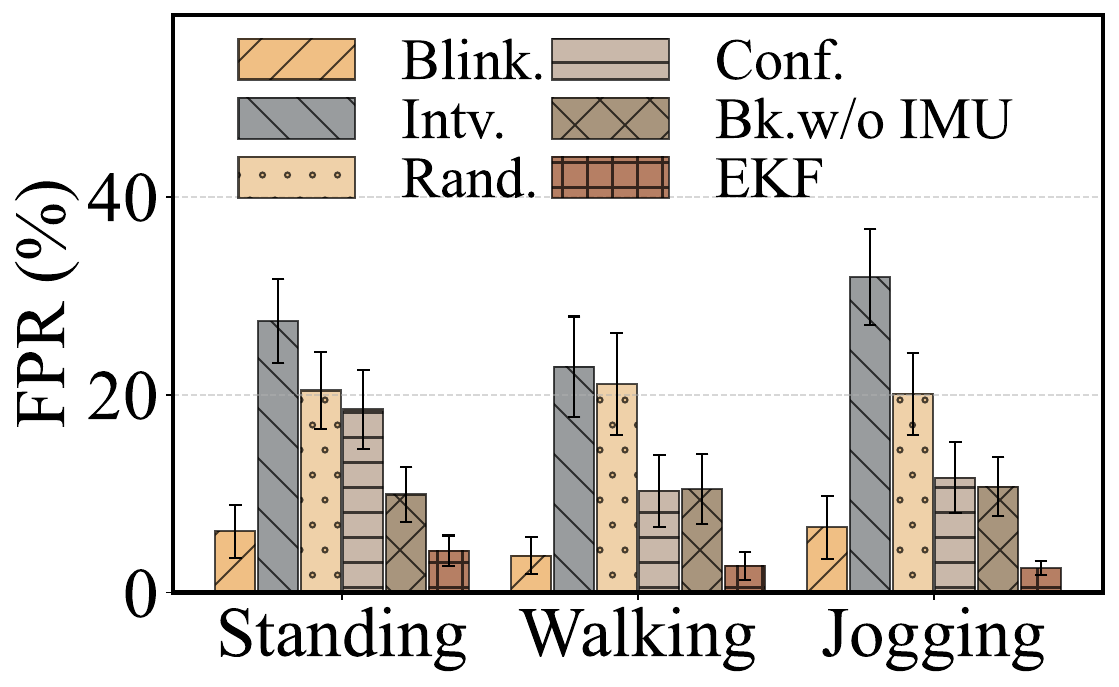} 
		\end{minipage}
		\label{fig:static}
	}
	\subfigure[FNR in 3 trans. modes]{
		\begin{minipage}[]{0.46\linewidth} 
			\centering
			\includegraphics[height=3cm]{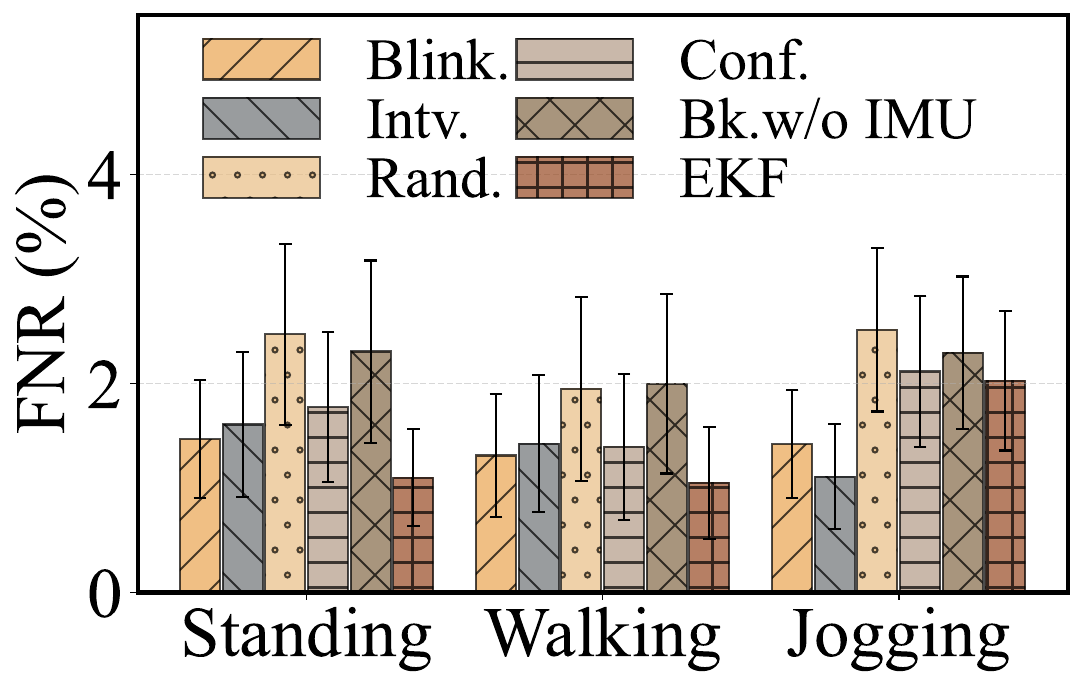} 
		\end{minipage}
		\label{fig:walk}
	}
	\caption{
    Hazard detection accuracy of BlinkBud and other candidate methods across different user transportation modes (standing, walking, jogging). BlinkBud achieves consistently lower FPR and FNR other than EKF with high system overhead, demonstrating robustness to ego-motion by leveraging adaptive sampling and IMU-based state estimation.}
	\label{fig:accuracy1}
\end{figure}





\subsubsection{Evaluation Metrics}
To evaluate the hazard detection accuracy of the system, we focus on two key metrics, \emph{i.e.}, false positive ratio (FPR) and false negative ratio (FNR), which are defined as follows:
\begin{equation*}
	\text{FPR} = \frac{n_{\text{fp}}}{n_a}, \text{FNR} = \frac{n_\text{fn}}{n_a},
\end{equation*}
where $n_a$ denotes the times of risk assessments and $n_{\text{fp}}$ denotes the number of assessments where the system erroneously detected a danger, and $n_\text{fn}$ denotes the number of assessments where the system failed to alert for a danger.
We also considered power consumption, \emph{i.e.}, the amount of power consumed per unit time.

\subsection{Hazard Detection Performance}

We investigate the detection accuracy of BlinkBud and other candidates on all 262 traces in various using cases, \emph{i.e}, user transportation mode, location, vehicle types, vehicle numbers, light conditions. For each trace, the data from the first minute is used to initialize the Q-table, while the remaining data is utilized for evaluation.

\begin{figure}[] 
	\centering
    \subfigure[FPR in two road types]{
		\begin{minipage}[]{0.46\linewidth} 
			\centering
			\includegraphics[height=3cm]{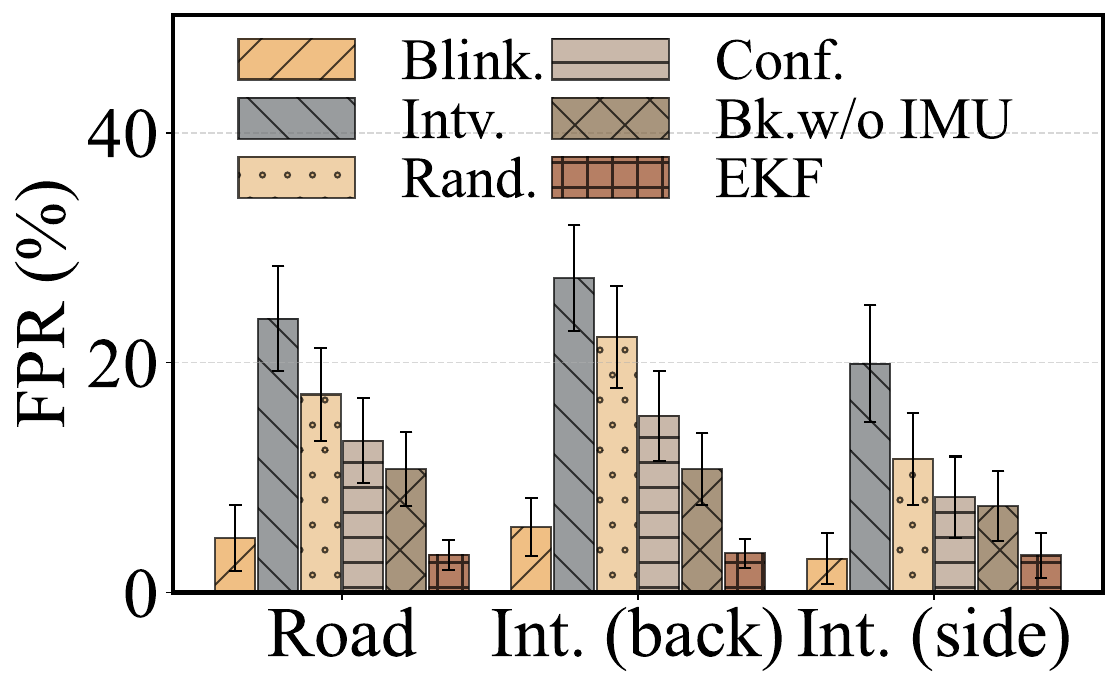} 
		\end{minipage}
		\label{fig:bridge}
	}
    \subfigure[FNR in two road types]{
		\begin{minipage}[]{0.46\linewidth} 
			\centering
			\includegraphics[height=3cm]{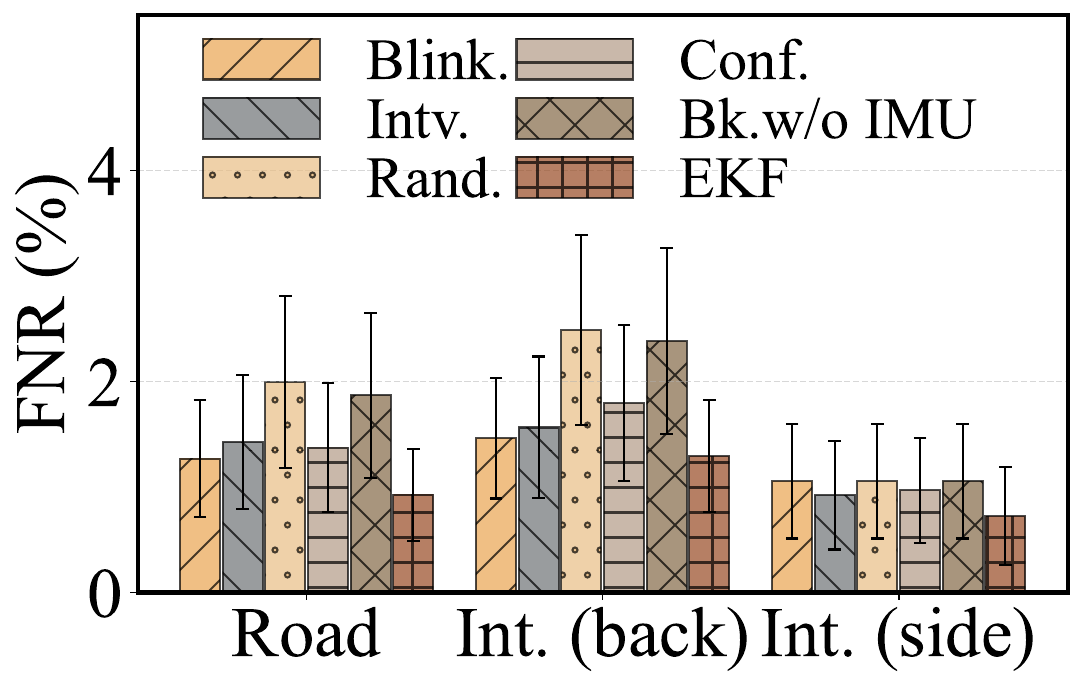} 
		\end{minipage}
		\label{fig:gateway}
	}
	\caption{Hazard detection accuracy under different road scenarios, including open roads and intersections with hazards approaching from behind, \emph{i.e.}, Int. (back), or the side, \emph{i.e.}, Int. (side). BlinkBud outperforms candidate methods with significantly reduced FPR and FNR other than EKF with high system overhead, particularly in side-approaching cases, highlighting the benefit of integrating motion awareness and dynamic sampling.}
	\label{fig:accuracy2}
\end{figure}

\begin{figure}[]
	\centering
	\subfigure[FPR in two vehicle types]{
		\begin{minipage}[]{0.46\linewidth}
			\centering
			\includegraphics[height=3cm]{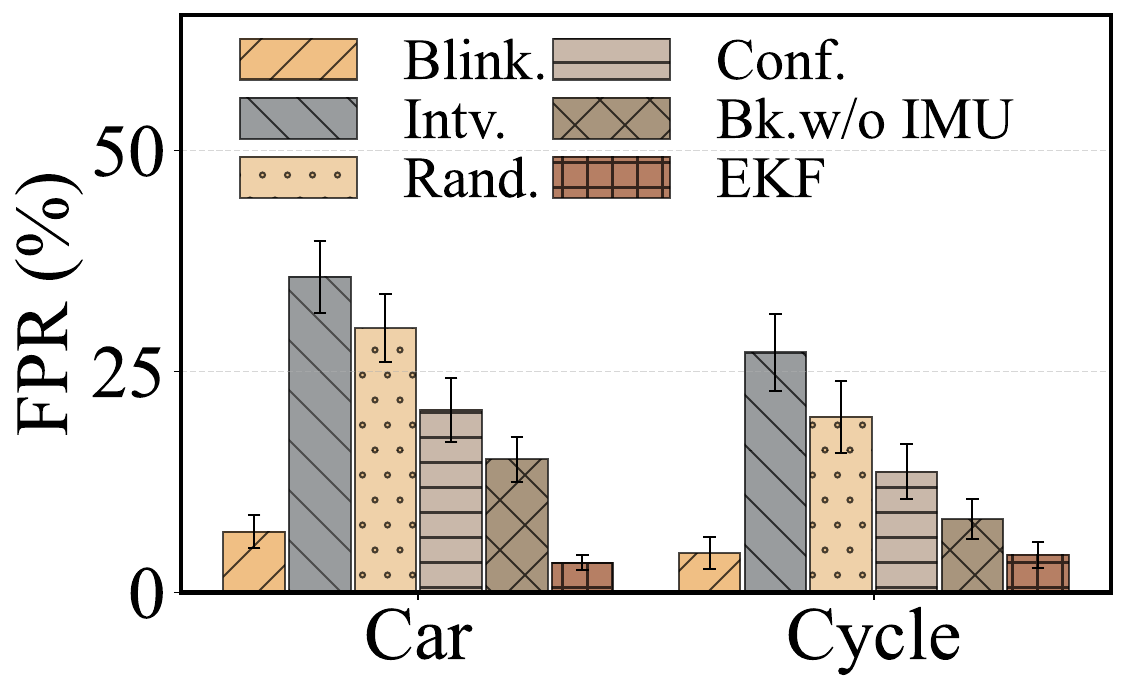}
		\end{minipage}
		\label{fig:largevehicle}
	}
	\subfigure[FNR in two vehicle types]{
		\begin{minipage}[]{0.46\linewidth}
			\centering
			\includegraphics[height=3cm]{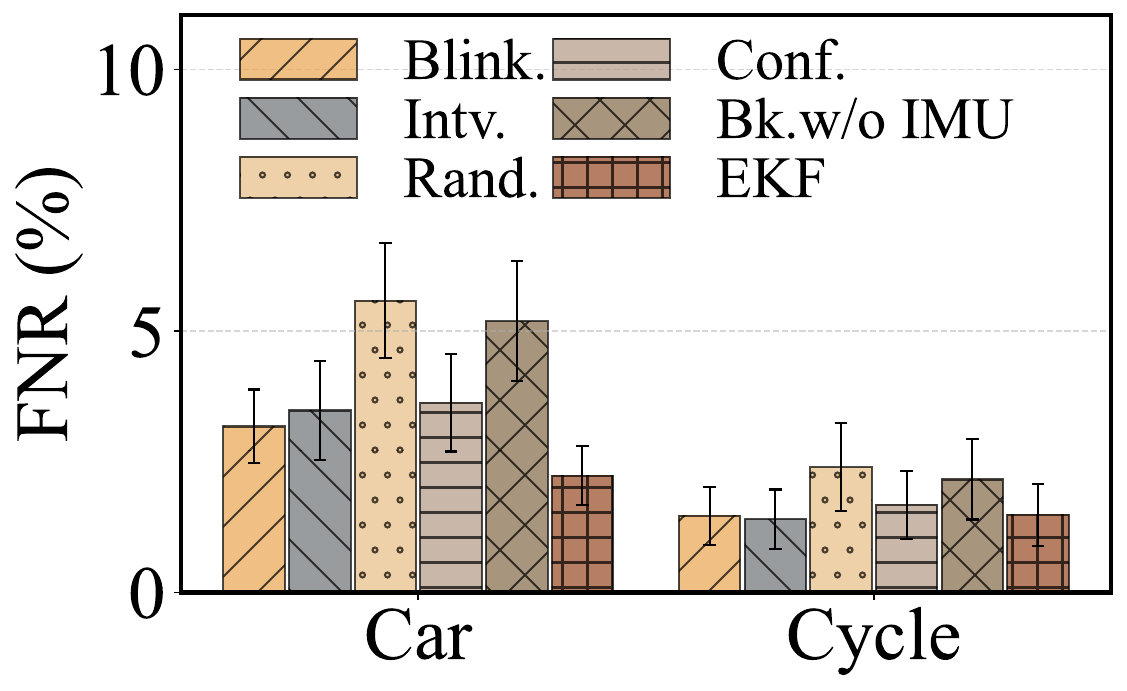}
		\end{minipage}
		\label{fig:smallvehicle}
	}
    \caption{Hazard detection accuracy of BlinkBud and other candidate methods on different vehicle types, where \emph{cycle} contains bicycle, e-cycle and motorcycle. BlinkBud consistently achieves lower FPR and FNR compared to other candidate methods other than EKF with high system overhead, demonstrating robustness across vehicle categories.}
    \label{fig:vehicle_sizes}
\end{figure}

\subsubsection{Impact of User Transportation Modes}

Figure \ref{fig:static}, Figure \ref{fig:walk} plot the FPR and FNR of different methods with different user transportation modes, \emph{i.e.}, static, walking, and jogging, respectively. We can see that the detection accuracy of BlinkBud surpasses all other methods. For example, in the walking mode, the FPR of BlinkBud can be reduced by up to 90.9\%. We can also see that BlinkBud has the highest accuracy in the walking state, followed by the static state. This is because, in the static state, although the body is still, the head tends to have more small movements, such as nodding and shaking, which makes detection more difficult. In the walking state, the large range of body movements and the frequent changes in pitch and yaw angles also pose challenges for warning accuracy. When the user runs, the camera will shake more violently, making the detection of danger more difficult. However, BlinkBud can maintain stable detection capability, with an FPR of only 6.6\% and an FNR of 1.4\%. This reflects the stability of BlinkBud's performance under intense motion.

\subsubsection{Impact of Road Tpyes}

Figure \ref{fig:bridge}, Figure \ref{fig:gateway} plot the FPR and FNR of different methods on different types of roads. On a road, vehicles can only arrive from behind. At the intersection, vehicles can arrive from both behind and the side. We can see that BlinkBud performs well in all two locations. This demonstrates the versatility of BlinkBud across user locations. Moreover, we can see that even if the vehicles come from the side, BlinkBud can accurately detect the hazards. This is because although the camera is limited to capturing vehicles positioned behind the user, the natural tendency of users to turn their heads when crossing the road enables intermittent visibility of approaching vehicles. Consequently, the tracking mechanism, supplemented by IMU-based corrections, can effectively monitor these vehicles despite their occasional appearance within the camera's field of view.

\subsubsection{Impact of Vehicle Types}

Figure \ref{fig:largevehicle} and Figure \ref{fig:smallvehicle} plot the FPR and FNR with cars and cycles. We note that in our collected dataset of rear-approaching objects, the only observed risks to pedestrians are cars and cycles, where cycle encompasses bicycle, e-cycle, and motorcycle. We can see that BlinkBud performs better in terms of alerting of cycles. This is because bicycles move relatively slowly and only need to be alerted when they are very close, which reduces the difficulty of alerting. The alerting FPR and FNR of cars is also less than 0.01, showing a high practicability of BlinkBud.

\begin{figure}[]
	\centering
	\subfigure[FPR in three vehicle numbers]{
		\begin{minipage}[]{0.46\linewidth}
			\centering
			\includegraphics[height=3cm]{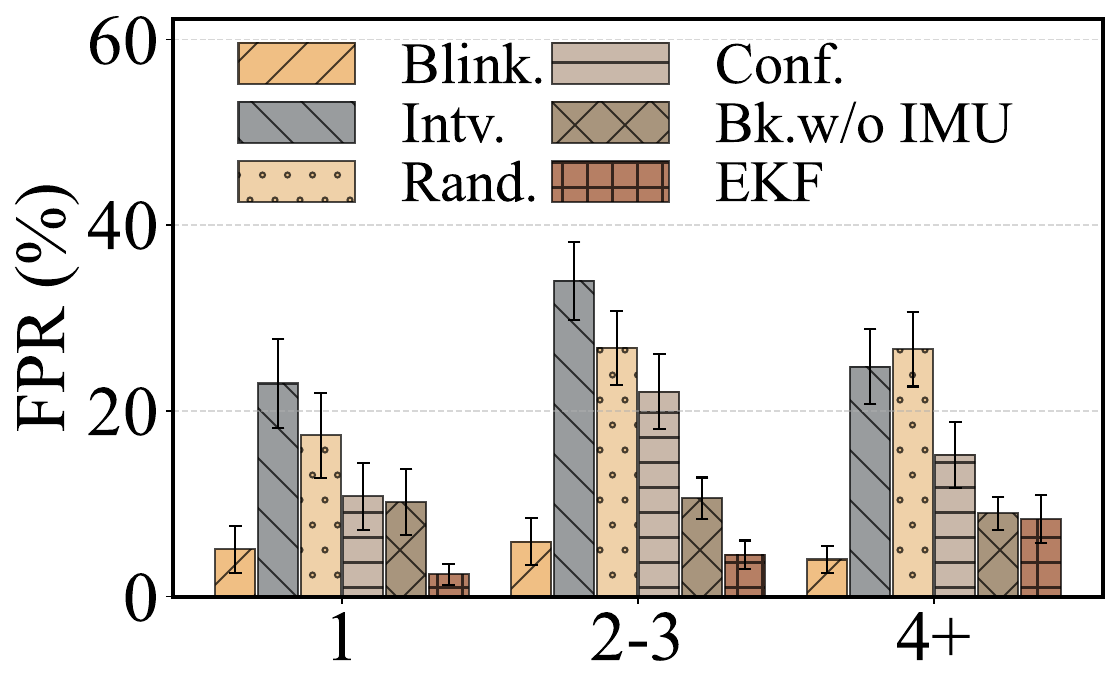}
		\end{minipage}
		\label{fig:onevehicle}
	}
	\subfigure[FNR in three vehicle numbers]{
		\begin{minipage}[]{0.46\linewidth}
			\centering
			\includegraphics[height=3cm]{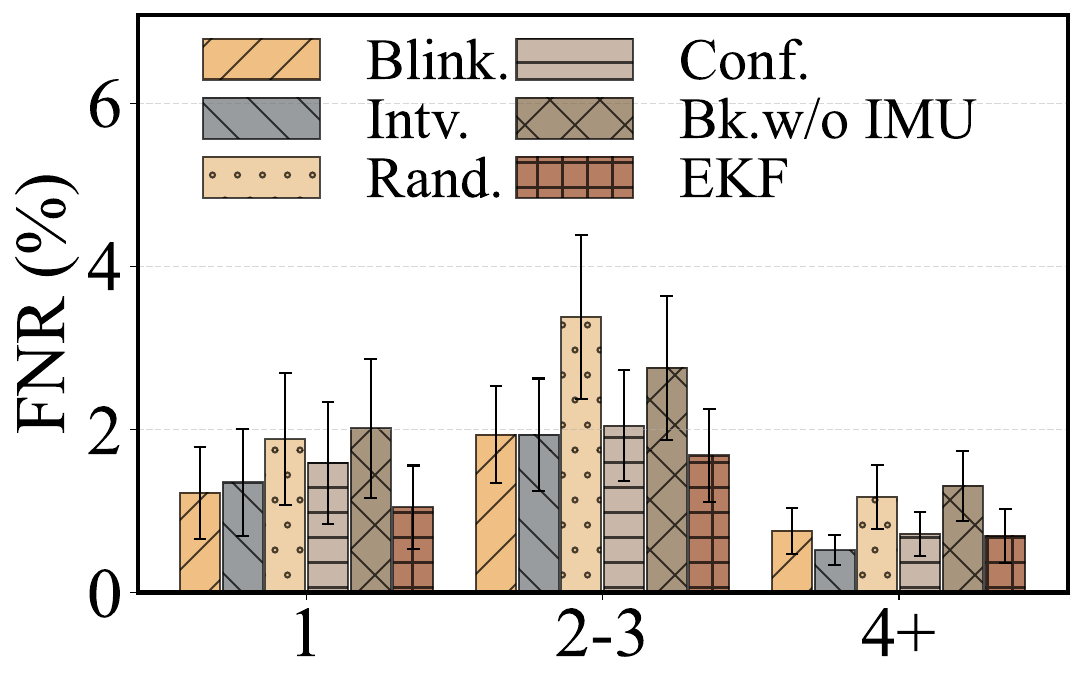}
		\end{minipage}
		\label{fig:multiplevehicle}
	}
    \caption{Hazard detection accuracy under different traffic densities, categorized by the number of simultaneously approaching vehicles. BlinkBud maintains consistently low FPR and FNR even in dense traffic (2–3 and 4+ vehicles), highlighting its robustness in crowded scenarios.}
    \label{fig:number_of_vehicles2}
\end{figure}

\begin{figure}[]
	\centering
	\subfigure[FPR in two light conditions]{
		\begin{minipage}[]{0.46\linewidth}
			\centering
			\includegraphics[height=3cm]{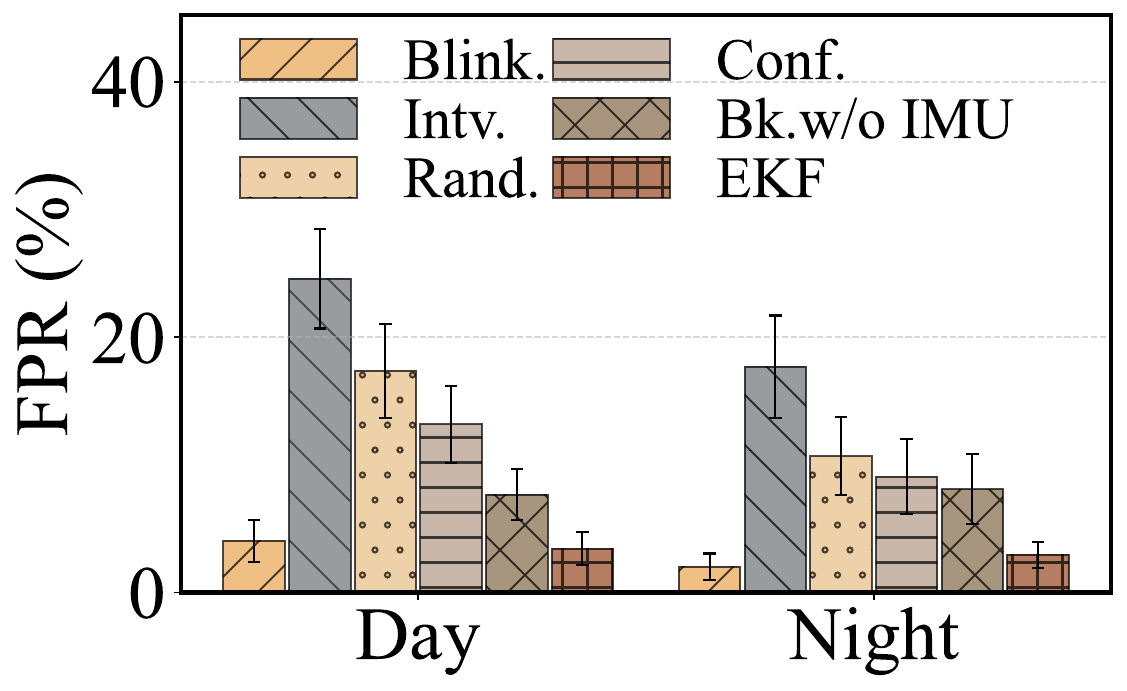}
		\end{minipage}
		\label{fig:day}
	}
	\subfigure[FNR in two light conditions]{
		\begin{minipage}[]{0.46\linewidth}
			\centering
			\includegraphics[height=3cm]{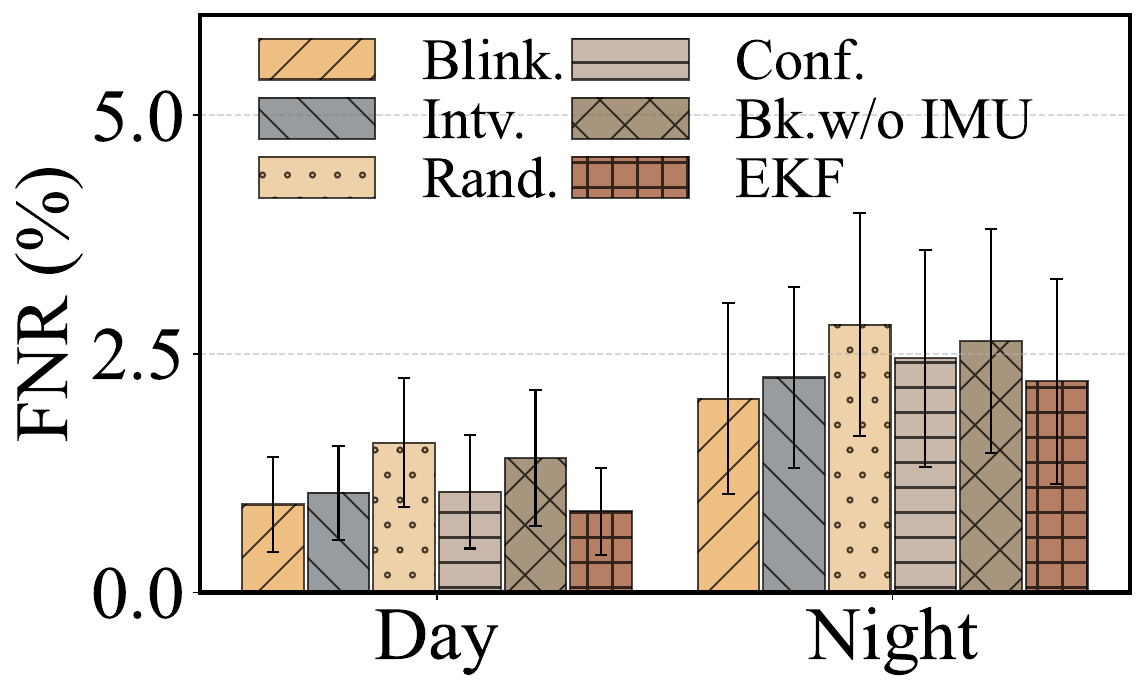}
		\end{minipage}
		\label{fig:night}
	}
	\caption{Hazard detection accuracy under varying light conditions, including day and night scenarios. BlinkBud demonstrates stable performance with low FPR and FNR across lighting variations, suggesting its resilience to illumination changes.}
	\label{fig:number_of_vehicles}
\end{figure}

\subsubsection{Impact of Vehicle Numbers}

Figure \ref{fig:onevehicle} and Figure \ref{fig:multiplevehicle} plot the FPR and FNR of detection accuracy with different numbers of vehicles. We manually labeled the number of vehicles included in each trace. We can see that when the vehicle number is only 1, BlinkBud performs very well, with both the FPR and FNR less than 0.01. When more vehicles arrive, BlinkBud also shows significant improvement compared to other methods, with almost a reduction of FPR by up to 94.74\%. The fixed interval sampling, \emph{i.e.}, Intv., cannot update the motion states of objects in time, leading to a significant increase in alerting FPR. We can also see surprisingly that the alerting error rate decreases when vehicle number exceeds 4. This is because the roads where people and vehicles mix are generally small, the roads are more crowded when vehicle number exceeds 4 and the driving speeds of vehicles become slower, causing a drop on the error rate.

%

\subsubsection{Impact of Light Conditions}

Figure \ref{fig:day} and Figure \ref{fig:night} plot the FPR and FNR with different light conditions. We can see that BlinkBud performs well both in day and night. The FNR of all methods during nighttime is slightly higher than that during the daytime because of the inherent defects of visual methods in low-lighting conditions. However, we can see that the FPR and FNR of BlinkBud are still low enough for use in night. This is because there are street lights at night, and the vehicles turn on the lights at night, so that the lighting conditions are generally not so bad that it is completely invisible.

\subsection{Overall Performance} 

\begin{figure}[]
	\centering
    \subfigure[FPR in all scenarios]{
		\begin{minipage}[]{0.46\linewidth} 
			\centering
			\includegraphics[height=3cm]{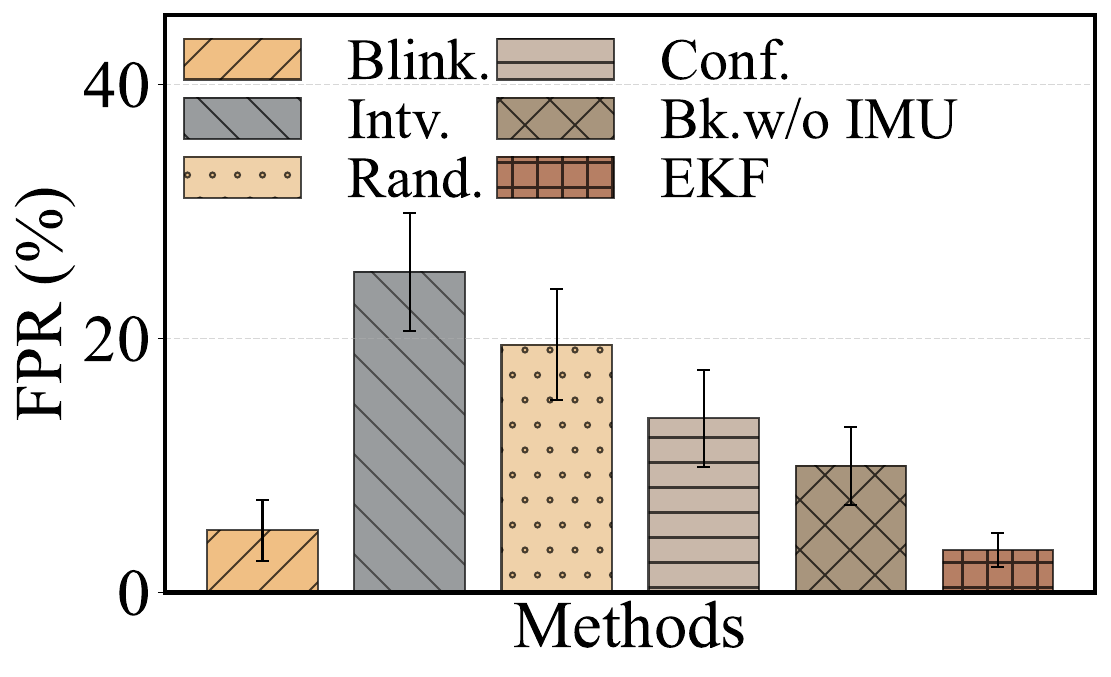} 
		\end{minipage}
		\label{fig:all_accurancy}
	}
    \subfigure[FNR in all scenarios]{
		\begin{minipage}[]{0.46\linewidth} 
			\centering
			\includegraphics[height=3cm]{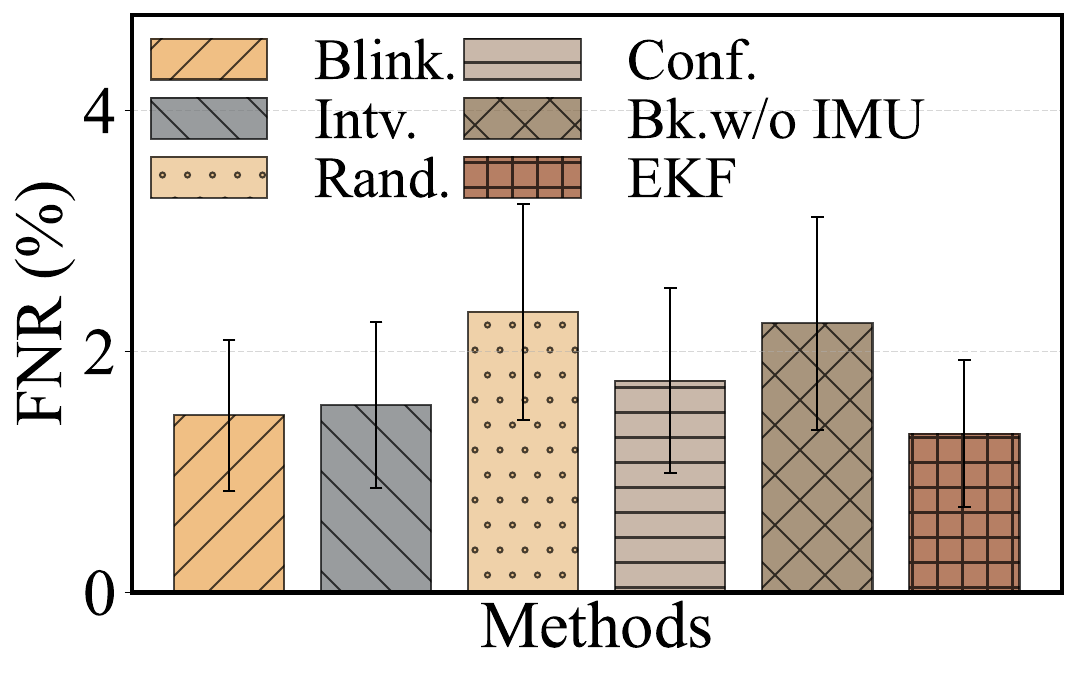} 
		\end{minipage}
		\label{fig:power}
	}
	\caption{Overall hazard detection accuracy across all evaluated scenarios. BlinkBud achieves the lowest average FPR (4.90\%) and FNR (1.47\%) among all candidate methods, demonstrating its consistent effectiveness in diverse real-world conditions.}
	\label{fig:overall}
\end{figure}

\subsubsection{Hazard Detection Accuracy}  

Figure \ref{fig:overall} plots the overall alerting FPR and FNR. BlinkBud demonstrate superior performance compared to all other candidate methods, especially on FPR. BlinkBud can reduce the FPR by up to 80.61\%. This is because Blinkbuses can sample at the right time to capture the object's motion state changes. We can also see that BlinkBud can achieve an FPR of 4.90\% and FNR of 1.47\%, showing its performance to detect hazards in time accurately. The detecting FPR is generally larger than FNR, especially for the Intv. method, whose FPR is 15.8$\times$ larger than FNR. This is because without sampling, Kalman filter will estimate the trajectory according to the current motion state of the object, ignoring the active avoidance measures that the driver may take, resulting in a large number of false positives. BlinkBud can actively learn the interval of vehicle motion state changes in different situations, and sample only when the vehicle motion state just changes, so as to achieve a low FPR of 4.90\%.

\begin{figure}[]
	\centering
	\subfigure[Earbud]{
		\begin{minipage}[]{0.48\linewidth} 
			\centering
			\includegraphics[height=3cm]{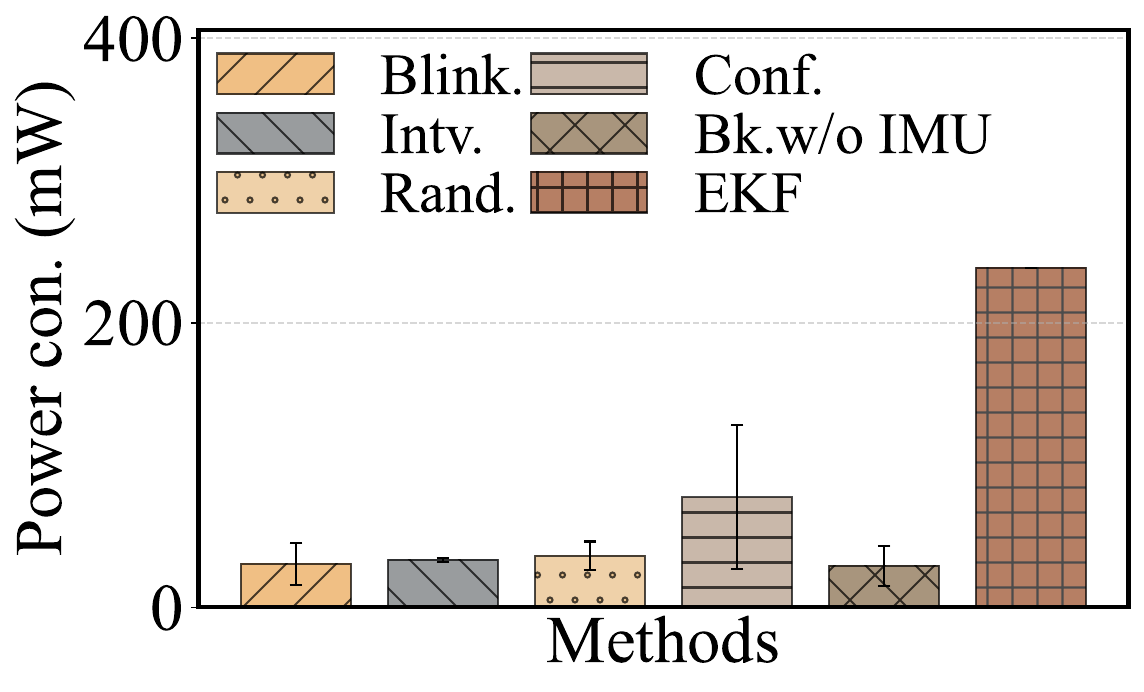} 
		\end{minipage}
		\label{fig:Power}
	}
	\subfigure[Smartphone]{
		\begin{minipage}[]{0.455\linewidth} 
			\centering
			\includegraphics[height=3cm]{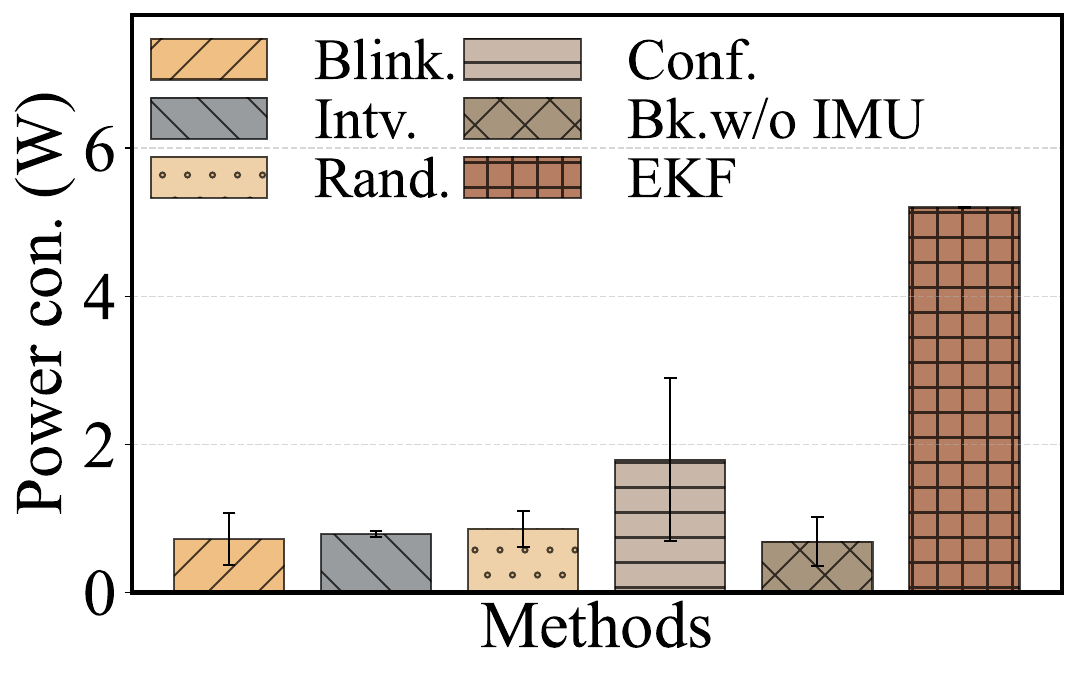} 
		\end{minipage}
		\label{fig:Power_phone}
	}
	\caption{Power consumption of earbud and phone of BlinkBud and other candidate methods, where BlinkBud has the lowest power consumption while maintaining high hazard detection accuracy.}
	\label{fig:overall_power}
\end{figure}

\begin{figure}[]
	\centering
	
	\includegraphics[width=0.5\linewidth]{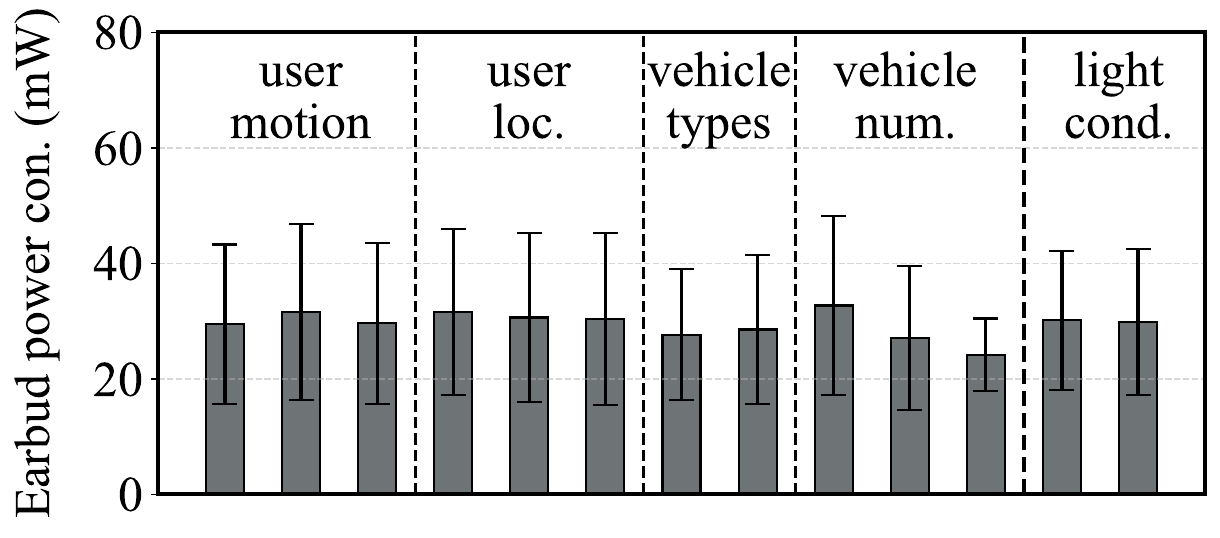} 
	\includegraphics[width=0.5\linewidth]
	{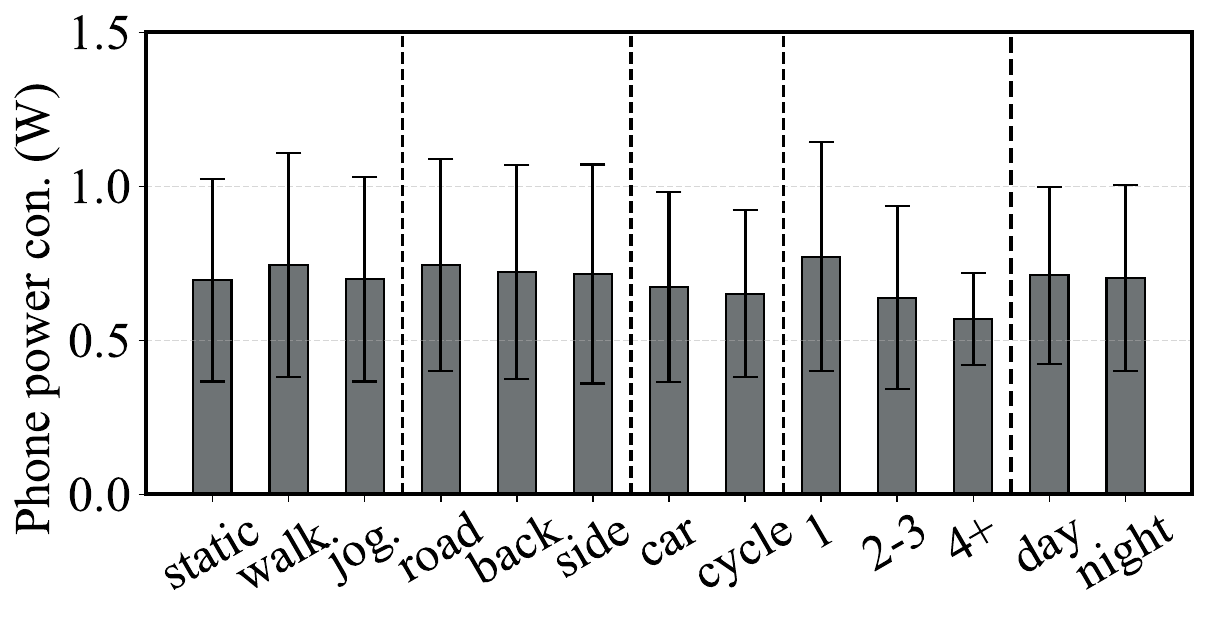}
	\caption{Power consumption of earbud and smartphone in different scenarios, where the average power on earbud and smartphone are 29.8 mW and 702.6 mW, with a battery life of 4.97 hours and 24.7 hours on the 40 mAh earbud battery and Mi11, respectively.}
	\label{fig:consumption_all}
\end{figure}

\subsubsection{Power Consumption}

We then investigate the power consumption of BlinkBud on both earbuds and smartphones. The power consumption of the earbud is measured using a multimeter. The power consumption of smartphone (represented by Mi 11) is measured using the power analyzer of Android Studio. 
Figure \ref{fig:overall_power} plot the power consumption of different methods on the earbud and smartphone. We can see that Blink can achieve close detection accuracy with a power consumption of only 13.87\% compared with the naive EKF methods. The detection accuracy of all methods whose power consumption is close to that of BlinkBud is much lower than that of BlinkBud. This indicates the effectiveness of BlinkBud's optimal blink sampling algorithm.

\begin{figure}[]
	\centering
	\subfigure[CPU usage]{
		\begin{minipage}[t]{0.46\linewidth}
			\centering
			\includegraphics[height=4.1cm]{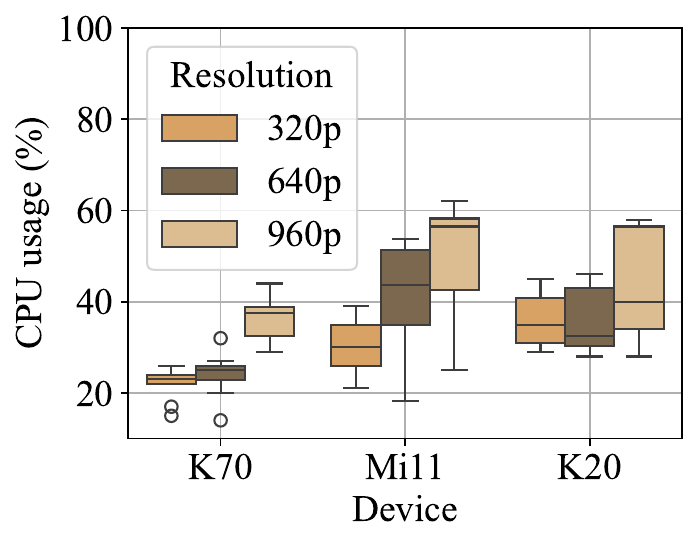}
		\end{minipage}
		\label{fig:cpuUsage}
	}
	\subfigure[Memory usage]{
		\begin{minipage}[t]{0.46\linewidth}
			\centering
			\includegraphics[height=4cm]{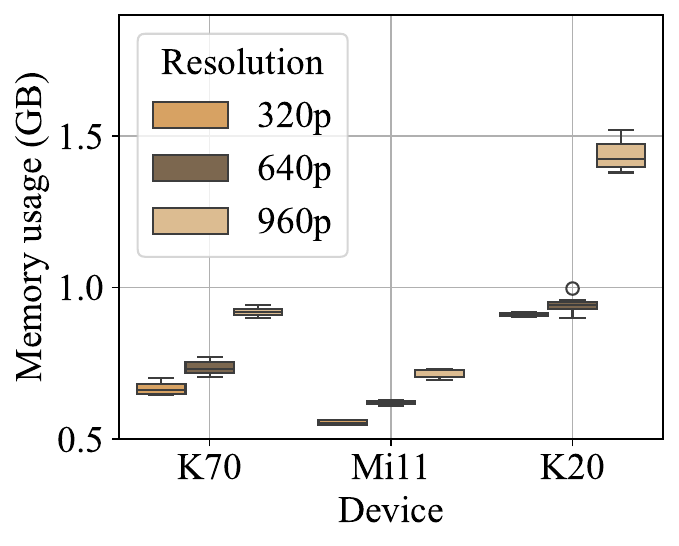}
		\end{minipage}
		\label{fig:memory}
	}
	\caption{CPU and memory usage of smartphones, where the overhead of BlinkBud is low for both low-end and high-end smartphones.}
	\label{fig:phone_overhead}
\end{figure}

We further plot the power consumption of BlinkBud on both earbud and smartphone over different scenarios in Figure \ref{fig:consumption_all}. 
In complex scenarios, such as high vehicle densities, the power consumption increases by around 22\%, accordingly, due to the need for more frequent sampling. Overall, the average BlinkBud power consumption remains around 29.8 mW, and the average smartphone power consumption remains around 702.6 mW.
Consider a battery of 40 mAh on the earbud, BlinkBud can continuously run for 4.97 hours. 
Such power consumption is acceptable, ensuring that our system has sufficient battery life for extended use.

\subsubsection{CPU and Memory Consumption}

We then investigate the CPU usage and memory consumption of the proposed BlinkBud system. Specifically, the phone’s CPU usage and memory usage are measured using the profiler in Android Studio.
All measured metrics are tested 30 times. The boxplots of CPU and memory usage are shown in Figure \ref{fig:phone_overhead}.
We can see from Figure \ref{fig:phone_overhead} that BlinkBud is lightweight, with a CPU usage of less than 50\% and memory usage of less than 1GB on average. This shows that the system load of BlinkBud is acceptable for existing smartphones.



\begin{figure}[]
	\centering
	\subfigure[BlinkBud can continuously track objects while $\theta^y$ changes, while V.O. has an incorrect estimation of the horizontal axis.]{
		\begin{minipage}[]{\linewidth}
			\centering
			\includegraphics[width=0.6\linewidth]{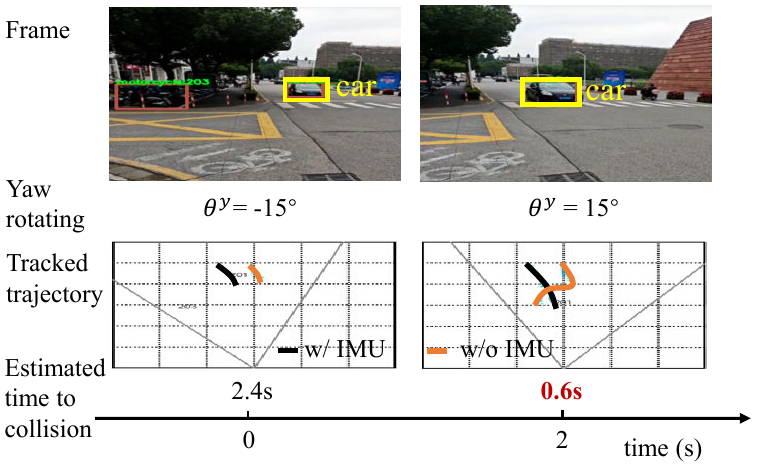}
		\end{minipage}
		\label{fig:yaw}
	}
	\subfigure[BlinkBud can continuously track objects while $\theta^p$ changes, while V.O. has an incorrect estimation of the object depth.]{
		\begin{minipage}[]{\linewidth}
			\centering
			\includegraphics[width=0.6\linewidth]{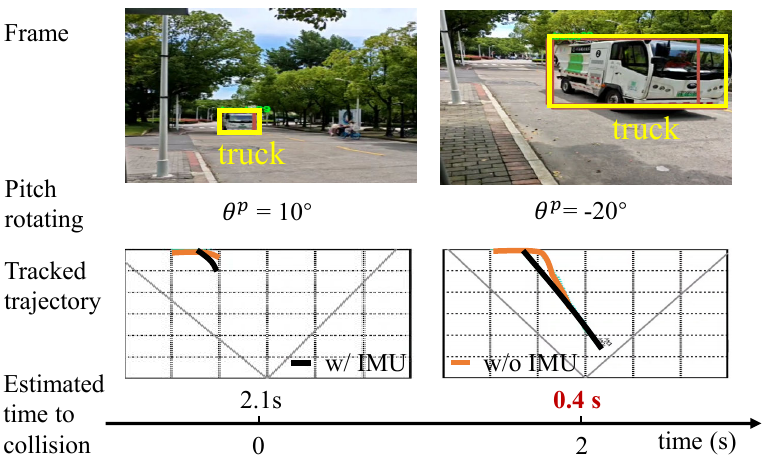}
		\end{minipage}
		\label{fig:pitch}
	}
	\caption{Visualization of rear object tracking results when the head is moving intensely with yaw and pitch rotations, respectively, where the tracking results are stable in both cases and alerts are successfully issued.}
	\label{fig:yaw_and_pitch}
\end{figure}

\begin{table}[]
\centering
\caption{User experience evaluation across five key dimensions using a 5-point Likert scale (1 = strongly disagree, 5 = strongly agree). The table reports the median rating and the interquartile range (IQR), which represents the spread between the 25th and 75th percentiles, reflecting the consistency of user responses.}
\label{tab:userexperience}
\scalebox{1}{
\begin{tabular}{ccc}
\toprule
Statement                                                           & Rating (Median) & IQR \\ \hline 
I am \textbf{likely} to use this system regularly.                           & 5   &   1      \\ 
The system is \textbf{effective} in detecting hazards.                 & 5    &    1    \\  
The system responds \textbf{quickly} to detected hazards.                      & 4     &   1    \\  
I am not concerned about my \textbf{privacy} when using the system.                     & 4     &  1     \\  
The system is efficient in terms of \textbf{power} consumption.                       & 4      &   1   \\ \bottomrule
\end{tabular}
}
\end{table}

\subsection{Visualization of Tracking Results}

To further illustrate the tracking results of BlinkBud, we illustrate two examples of tracking when the camera is moving intensely in the yaw and pitch directions, respectively. Specifically, we designed an experiment to test whether the change of headphone pose affects the system's hazardous objects tracking and danger alerting ability. The volunteers rotated the head in yaw and pitch directions during the test to collect the videos with the shaking of the earbuds when a person was walking. 

The visualization results are shown in Figure \ref{fig:yaw_and_pitch}. A car and a truck are tracked in different poses in chronological order. We can see that in this case, the change of camera pose does not affect the accuracy of detection and tracking as well as danger assessment. This demonstrates the superior performance of BlinkBud in real-world scenarios with intense head movements. More examples of videos are shown in this anonymous \href{https://www.dropbox.com/scl/fo/bezlwf9p2hpkrqyhd25ew/AH38fzNC9JMf3ooUQtoP4bc?rlkey=rntynugs7svmdtop9som859v9&st=tq64g6su&dl=0}{\color{cyan} link}.

\subsection{User Experience Survey}

We conducted a formative user experience survey. A total of 43 volunteers were recruited to use the BlinkBud system for a day. The volunteers ranged in age from 18 to 32 years old, with 31 males and 12 females. Their heights spanned from 1.6m to 1.9m. The volunteers included individuals with different levels of familiarity with wearable technology, from first-time users to regular users.
They rated it from five perspectives: willingness (\emph{i.e.}, is this type of application acceptable), effectiveness (\emph{i.e.}, is the functionality useful), response time (\emph{i.e.}, is the system smooth enough), privacy concern (\emph{i.e.}, does it effectively protect the privacy of users and others), and power consumption (\emph{i.e.}, whether concerned about excessive power consumption of the earphones and phone). The specific statements and results are shown in Table \ref{tab:userexperience}. Note that the questionnaire is intended to capture users’ preliminary subjective impressions, reflecting practical usability and psychological acceptance, rather than providing conclusive evidence.

As shown in Table \ref{tab:userexperience}, users’ responses provide formative insights into BlinkBud’s usability and acceptance. While most participants generally perceived the system as useful and acceptable, some expressed mild concerns regarding system latency, privacy, and power consumption. Although the average scores are relatively high, the interquartile ranges (IQRs) for all aspects of BlinkBud indicate noticeable variability in user responses. This is because users have different levels of concern for their own safety and privacy, leading to varied user experiences.

\section{LIMITATIONS AND FUTURE WORK}

\subsection{Limitations of the Formative User Study}


Our user study is inherently formative rather than conclusive. It was conducted over a single day and relied solely on self-reported questionnaires, without incorporating task-based usability trials, objective behavioral measurements, or validated assessment instruments such as the System Usability Scale (SUS) or NASA-TLX. Consequently, the findings primarily provide preliminary insights into user perceptions, subjective workload, and perceived usability, but they should not be interpreted as definitive evidence of the system’s overall performance or effectiveness. Moreover, the absence of repeated measures or longitudinal observation limits our ability to assess learning effects, sustained engagement, or real-world usage patterns. To address these limitations, future work will involve more rigorous experimental designs, including controlled task-based studies, objective behavioral metrics, and standardized usability and workload assessments. Such approaches will enable more conclusive, reliable, and generalizable evaluations of both user experience and system performance across diverse participant populations and real-world conditions.

\subsection{Limitations of External Validity}

The external validity of our study is indeed limited. Our evaluation was conducted with only ten participants and restricted to a single university campus under limited weather and traffic conditions. These constraints may affect the generalizability of the findings to broader real-world scenarios. Nevertheless, we note that the campus environment represents a prototypical mixed pedestrian-vehicle setting, where many of the key challenges encountered in urban mobility, such as blind spots, shared space negotiation, and variable pedestrian behaviors, are naturally present. This provides at least partial external validity despite the limited scope of the study. To strengthen generalizability, future work will extend the evaluation to diverse environments, including urban streets, residential neighborhoods, and parking lots, as well as under different traffic densities and weather conditions. Such extensions will allow us to assess the robustness of the system across a broader range of contexts and to validate its usability and effectiveness in more representative real-world deployments.

\subsection{Limitations in Low-light Conditions}

As our object tracking is vision-based, detection accuracy will decrease in very low-light environments when the camera fails to capture clear images, and the potential risk may not be alarmed promptly.
Results of our experiments show that even at night, the tracking accuracy will not encounter significant losses if there are streetlights or vehicle headlights nearby. Therefore, BlinkBud works under most circumstances. For minor nighttime conditions with insufficient ambient light, we are considering adding a notification feature to the earbuds. This feature will remind the user to be cautious when the camera detects low image brightness that may impair effective hazard detection.

\section{CONCLUSION}

In this paper, we present BlinkBud, an innovative system that leverages real-time images captured by a miniature rear-facing camera integrated into earbuds to detect and track hazardous objects approaching from behind. By incorporating a lightweight monocular 3D object tracking algorithm and depth estimation techniques, along with an optimal blink sampling decision algorithm, BlinkBud effectively identifies and alerts pedestrians to potential threats posed by oncoming vehicles. BlinkBud is lightweight and computationally efficient which ensures seamless deployment on everyday mobile devices, while its integration of gyroscopic data enables robust trajectory estimation that is independent of head movements. BlinkBud provides a practical and computationally efficient approach for mitigating risks associated with rear blind spots in real-world pedestrian scenarios, and demonstrates potential for enhancing user situational awareness and safety.

\section*{Acknowledgments}
This work was supported in part by the Natural Science Foundation of China (Grants No. 62432008, U21A20462, 62472083 and 62572098). We also thank Mr. Jinxu Zhou from Nanjing University of Aeronautics and Astronautics for his support in 3D printing.



\bibliographystyle{ieeetr}
\bibliography{sample-base}

\appendix
\section{Depth Estimation}
\label{sec:depth}
For each $o^{\mathbb{I}}$ in the detected 2D objects $\{o^{\mathbb{I}}_1, o^{\mathbb{I}}_2, \cdots, o^{\mathbb{I}}_{n} \}$, the depth $z^{\mathbb{C}}$ of $o^{\mathbb{I}}$ is then estimated using the camera’s intrinsic parameters and the pitch angle provided by the gyroscope. The core idea is to calculate depth based on the vertical position of a vehicle’s projection relative to the horizon in the image plane. The horizon serves as a reference line, and depth is determined using the pixel deviation between the vehicle’s bottom contact point with the ground and the horizon.

Specifically, we first calculate the pitch angle $\theta^{p}$ of the user's head using the IMU data, which represents the tilte of the user's head relative to the horizon. Then, the horizon $y_h^{\mathbb{I}}$ in the image coordinate system $\mathbb{I}$ can be calculated:
\begin{equation*}
    y_h^{\mathbb{I}} = c_y - f_y \tan \theta^p,
\end{equation*}
where $c_y$ is the vertical coordinate of the camera’s principal point in the image, and $f_y$ is the focal length along the vertical axis in pixel units. Both $c_y$ and $f_y$ are derived from the camera's intrinsic parameters.
The vertical pixel deviation between the contact point of object $o^{\mathbb{I}}$ and the horizon $y_h^{\mathbb{I}}$, denoted as $\Delta y^{\mathbb{I}}$, is then computed as:
\begin{equation*}
    \Delta y^{\mathbb{I}} = y^{\mathbb{I}} + h^{\mathbb{I}} - y_{h}^{\mathbb{I}},
\end{equation*}
where $y^{\mathbb{I}}$ is the vertical coordinate of the object’s bottom edge, and $h^{\mathbb{I}}$ is its height in the image plane. Finally, the depth $z^{\mathbb{C}}$ of vehicle $v$ is calculated based on the Pinhole projection model as follows:
\begin{equation*}
    z^{\mathbb{C}} = \frac{f_y h_e}{\Delta y^{\mathbb{I}}},
\end{equation*}
where $h_e$ is the height of the camera above the ground. Note that since the camera is mounted on the earbud, which is positioned on the user’s body, $h_e$ can be estimated based on the user's height.




\section{2D to 3D Coordinate Transformation} 
\label{sec:trans}

Given each detected 2D object $o^{\mathbb{I}}$ in $o^{\mathbb{I}}_1, o^{\mathbb{I}}_2, \cdots, o^{\mathbb{I}}_{n}$ and its depth $z^{\mathbb{C}}$ in $z^{\mathbb{C}}_1, z^{\mathbb{C}}_2, \cdots, z^{\mathbb{C}}_n$, we transforms $o^{\mathbb{I}}$ to 3D objects in the camera coordinate using the measurable camera intrinsic parameter matrix $K^{i}$:
\begin{equation*}
    \begin{bmatrix}
    x^{\mathbb{C}} \\
    y^{\mathbb{C}} \\
    z^{\mathbb{C}}
    \end{bmatrix} = \frac{z^{\mathbb{C}}}{K^{i}} \begin{bmatrix}
    x^{\mathbb{I}} \\
    y^{\mathbb{I}} \\
    1
    \end{bmatrix},
\end{equation*}
where $x^{\mathbb{I}}$ and $y^{\mathbb{I}}$ denotes the value of two axes in the image coordinate system, respectively.

\section{Multi-object Tracking}
\label{sec:mot}

For each object $o^{\mathbb{C}}_t$ matched across frames at time step $t$, we first transform the coordinates from the camera coordinate system to the user coordinate system, considering the yaw angle $\theta^y$ between the two coordinate systems :
\begin{equation*}
    \begin{bmatrix}
    x^{\mathbb{U}}_t \\
    z^{\mathbb{U}}_t
    \end{bmatrix} = \begin{bmatrix}
    cos \theta^y_t & -sin \theta^y_t \\
    sin \theta^y_t & cos \theta^y_t
    \end{bmatrix} \begin{bmatrix}
    x^{\mathbb{C}}_t \\
    z^{\mathbb{C}}_t
    \end{bmatrix},
\end{equation*}
where $x^{\mathbb{C}}_t$ and $z^{\mathbb{C}}_t$ denotes the coordinates of x-axis and z-axis for object $o^{\mathbb{C}}_t$.
Then, we denote the states considered for tracking $X_t$ as follows:
\begin{equation*}
    X_t = \begin{bmatrix}
    x_t^{\mathbb{U}} &
    z_t^{\mathbb{U}} &
    \dot{x}_t^{\mathbb{U}} &
    \dot{z}_t^{\mathbb{U}}
    \end{bmatrix}^T,
\end{equation*}
where $\dot{x_t^{\mathbb{U}}}$ and $\dot{z_t^{\mathbb{U}}}$ denotes the velocity of object $o^{\mathbb{U}}_t$ in the x and z directions, respectively.
Considering that the motion of the vehicle between every two samples is a uniform linear motion, the motion function of object $o^{\mathbb{U}}_t$ is built as follows:
\begin{equation*}
    X_{t+1} = F_tX_t + w_t,
\end{equation*}
where $w_t$ is process noise and $F_t$ denotes the state transition matrix:
\begin{equation*}
    F_t = \begin{bmatrix}
    1 & 0 & \Delta t & 0 \\
    0 & 1 & 0 & \Delta t \\
    0 & 0 & 1 & 0 \\
    0 & 0 & 0 & 1
    \end{bmatrix}.
\end{equation*}
The observation $Y_t$ of object $o^{\mathbb{C}}_t$ is set as its coordinate values $x_t^{\mathbb{U}}$, $y_t^{\mathbb{U}}$ and height $h_t^{\mathbb{U}}$ in the 2D image coordinate system $\mathbb{U}$. The relationship between $Y_t$ and $X_t$ is as follows:
\begin{equation*}
\label{equ:update}
\bm{Y}_t = 
\left[ 
\begin{aligned}
& \frac{f_x \left( x_t^{\mathbb{U}} \cos \theta^{y}_t + z_t^{\mathbb{U}} \sin \theta^{y}_t \right)}{-x_t^{\mathbb{U}} \sin \theta^{y}_t + z_t^{\mathbb{U}} \cos \theta^{y}_t} \cos \theta^{p}_t \\
& \frac{f_y h_t^{\mathbb{U}}}{-x_t^{\mathbb{U}} \sin \theta^{y}_t + z_t^{\mathbb{U}} \cos \theta^{y}_t} \\
& \frac{f_y h_e}{-x_t^{\mathbb{U}} \sin \theta^{y}_t + z_t^{\mathbb{U}} \cos \theta^{y}_t}
\end{aligned}
\right] + w_t,
\end{equation*}
where $f_x$ denotes a constant of the camera’s horizontal focal length and $w_t$ denotes observation noise. Due to the non-linear relationship between $X_t$ and $Y_t$, the Extended Kalman Filter (EKF) is utilized to estimate the state of the target. The Extended Kalman Filter (EKF) first estimates the future state of the objects based on their current state and motion function, then corrects predictions based on observed results. This iterative process continues for each frame, ensuring accurate tracking of the target's position and velocity. If a target is not detected for multiple frames, it is removed from the tracking queue. In this way, a set of trajectories $r^{\mathbb{U}}_1, r^{\mathbb{U}}_2, \cdots, r^{\mathbb{U}}_{n}$ of the detected objects can be tracked.

\section{Optimal Blink Sampling Decision}
\label{sec:blink_sampling}

Using the camera on the earbuds for accurate tracking, a naive approach is to record all frames at the highest rate and transmit them to the smartphone for computing. However, this naive approach results in excessive power consumption and unnecessary high-frequency sampling, as hazards are not always present. To this end, we propose an adaptive frame sampling strategy based on an online reinforcement learning method, \emph{i.e.}, SARSA algorithm, to dynamically balance computational efficiency and detection accuracy. The core idea of the adaptive frame sampling strategy is to learn a value function for each state $S$ and each action $A$. At each time step, we determine the optimal action based on the current state and the value function and iteratively update the value function using rewards obtained from executing actions. Formally, we first reformulate the RHD problem as a Markov decision problem as follows:

\emph{1) State ($S$)}: The current state of the system, including the following three items:
\begin{itemize}
\item \emph{Object Lowest Confidence}: The lowest confidence of currently tracked objects, derived from the tracking based on EKF as the inverse of the trace of the covariance matrix. This reflects the system’s certainty regarding the accuracy of the object’s estimated position.
\item \emph{Distance of Object with Lowest Confidence}: The distance of the object from the user with lowest confidence, \emph{i.e.}, $z$, as closer objects are typically more critical for ensuring pedestrian safety. This factor prioritizes objects that pose an immediate hazard.
\item \emph{Time Since Last Sampling}: The time elapsed since the last frame was sampled, $\Delta t$. This ensures that new objects can be detected promptly, preventing blind spots caused by prolonged non-sampling periods.
\end{itemize}
These factors collectively define the system’s state space, allowing BlinkBud to dynamically adapt its frame sampling behavior to balance energy efficiency and detection accuracy.



\emph{2) Action ($A$)}: The action space consists of deciding whether to sample the environment ($A=1$) or not ($A=0$) at the current timestep.

\emph{3) Transition Function ($P(S_{t+1} | S_t, A_t)$)}: The system’s evolution is governed by the vehicle dynamics, sensor observations, and the tracking process, which updates the estimated state and confidence based on sampling.

\emph{4) Reward ($R(S, A)$)}: The reward function in the adaptive sampling algorithm is designed to balance the computational cost of sampling with the benefit of improved confidence in object tracking. At each time step $t$, the reward  R($S_t$, $A_t$)  reflects both the penalty for sampling and the improvement in confidence for tracked objects, encouraging efficient sampling strategies. Formally, the reward function is defined:
\begin{equation*}
    R(S_t, A_t) = A_t \cdot C_{s} + \Delta c,
\end{equation*}
where $\Delta c = c_{t+1} - c_t$  denotes the improvement in confidence, where $c_t$ is the confidence at time  $t$; $A_t \in \{0, 1\}$ is the action at time $t$, where $A_t = 1$ indicates sampling, and $A_t = 0$ indicates no sampling; $C_{s}$ represents the cost of sampling.

To solve this Markov decision problem, we first initialize a $Q$-value function  $Q(S, A)$ arbitrarily for all state-action pairs. At each time step $t$, BlinkBud observe the current state, denoted as $S_t$. 
Then, an action is selected using an  $\epsilon$-greedy policy:
\begin{equation*}
    A_t =
\begin{cases}
\text{Random Action} & \text{with probability } \epsilon, \\
\arg\max_{A} Q(S_t, A) & \text{with probability } 1-\epsilon.
\end{cases}
\end{equation*}
The probability decays as follows:
\begin{equation*}
    \epsilon = \frac{\epsilon_0}{1 + \eta t},
\end{equation*}
where $\epsilon_0 > 0$ represents the initial exploration rate, $\eta > 0$ controls the decay rate, and $t$ is the current time step. 
Action $A_t$ will be executed, and the resulting state $S_{t+1}$  and reward  $R_t$ can then be observed. The value function for the current state action pair can then be updated as follows:
\begin{equation*}
    Q(S_t, A_t) \gets Q(S_t, A_t) + \alpha \left[ R_t + \beta Q(S_{t+1}, A_{t+1}) - Q(S_t, A_t) \right],
\end{equation*}
where $A_{t+1}$ denotes the action selected in the next state $S_{t+1}$; $R_t$ denotes the reward of $S_t$ and $A_t$, \emph{i.e.}, $R_t = R(S_t, A_t)$; $\alpha$ denotes the learning rate that controls how quickly the value function is updated based on new information; $beta$ denotes discount factor that determines the importance of future rewards relative to immediate rewards. The learning rate $\alpha$ is set to gradually decrease to ensure convergence:
\begin{equation*}
    \alpha_t(s, a) = \frac{1}{N_t(s, a)},
\end{equation*}
where $N_t(s, a)$ denotes the number of times the state-action pair $(s, a)$ has been visited up to time step  $t$.
The process will be repeated for all time steps.

\section{Proof of Theorem \ref{theo}}
\label{sec:proof}

\begin{proof}

The adaptive frame sampling strategy is designed based on the SARSA algorithm. The asymptotic convergence of SARSA to the optimal policy is guaranteed, provided that the policies from the policy improvement operator satisfy the “greedy in the limit with infinite exploration (GLIE)” condition.
We then prove the adaptive sampling algorithm satisfies the GLIE condition. To this end, we verify the two components of GLIE.

\emph{1) Infinite Exploration.} To ensure infinite exploration, it must be shown that every state-action pair $(s, a)$ is visited infinitely often during the learning process. First, the algorithm uses an  $\epsilon$-greedy policy, where at each time step, an exploratory action is chosen with probability $\epsilon_t > 0$, and the greedy action is chosen with probability $1 - \epsilon_t$. Second, the exploration probability $\epsilon_t$ is designed to decay over time following the schedule $\epsilon_t = \frac{\epsilon_0}{1 + \eta t}$. This schedule ensures that $\epsilon_t \to 0$ as $t \to \infty$, since the denominator $1 + \alpha t$ grows unbounded. At the same time, the series $\sum_{t=1}^\infty \epsilon_t = \infty$, as $\epsilon_t$  asymptotically resembles the terms of a harmonic series. 
Since the state-action space is finite, and $\epsilon_t$-greedy exploration guarantees a non-zero probability of visiting any $(s, a)$ infinitely often, the infinite exploration condition is satisfied.

\emph{2) Greedy in the Limit.} To ensure that the policy converges to a greedy policy in the limit, we must show that as  $t \to \infty$, the exploratory component $\epsilon_t$ vanishes and the action selection becomes entirely greedy $\pi(s) = \arg\max_{a} Q(s, a)$. This follows directly from the decay of $\epsilon_t$, where  $\epsilon_t \to 0$ as $t \to \infty$. When $\epsilon_t$ becomes sufficiently small, the algorithm selects the greedy action $\arg\max_a Q(s, a)$ with probability approaching 1. Furthermore, the SARSA update rule ensures that the value function converges to a fixed point that satisfies the Bellman equation for the current policy. As $\epsilon_t \to 0$, the policy converges to a deterministic greedy policy for the optimal action-value function $Q^*(s, a)$.

In conclusion, the adaptive frame sampling strategy satisfies the GLIE condition and therefore converges to the optimal policy.
\end{proof}

\end{document}